\documentclass[twocolumn,10pt]{wlscirep}

\usepackage[utf8]{inputenc}
\usepackage[T1]{fontenc}
\usepackage{hyperref}

\usepackage{tabularx}
\usepackage{pifont}
\usepackage{wrapfig}
\usepackage{graphicx}
\usepackage{colortbl}
\usepackage{pgffor}
\usepackage{lipsum}
\usepackage{multirow}
\usepackage{makecell}
\usepackage{amsmath,amssymb}
\usepackage{bm}
\usepackage{tcolorbox}
\usepackage{listings}
\usepackage{float}
\usepackage[left, switch]{lineno}

\usepackage[capitalize]{cleveref}
\crefname{proposition}{Prop.}{Props.}
\crefname{section}{Sec.}{Secs.}
\crefname{table}{Tab.}{Tabs.}
\crefname{proposition}{Prop.}{Props.}
\crefname{section}{Sec.}{Secs.}
\crefname{table}{Tab.}{Tabs.}

\usepackage{algpseudocode}
\usepackage[ruled,vlined]{algorithm2e}

\usepackage{mwe}
\usepackage{cuted}
\usepackage{caption}

\makeatletter
\newcommand{\thickhline}{%
    \noalign {\ifnum 0=`}\fi \hrule height 1pt
    \futurelet \reserved@a \@xhline
}

\definecolor{mydarkyellow}{RGB}{216, 214, 196}
\definecolor{mymiddleyellow}{RGB}{229, 228, 218}
\definecolor{mylightyellow}{RGB}{245, 245, 240}
\definecolor{mygray}{gray}{.9}
\definecolor{mylightgray}{gray}{.95}
\definecolor{mygreen}{RGB}{93,173,85}
\definecolor{lightyellow}{RGB}{255, 255, 226}    
\definecolor{lightgray}{RGB}{248, 248, 248}     
\definecolor{lightpurple}{RGB}{247, 245, 235}     
\definecolor{darkgreen}{RGB}{55, 120, 46}
\definecolor{blue}{RGB}{51, 94, 150}
\definecolor{orange}{RGB}{220, 129, 79}
\definecolor{lightpink}{RGB}{253, 244, 241}
\definecolor{lightblue}{RGB}{245, 255, 255}
\definecolor{lightgreen}{RGB}{255, 255, 241}

\definecolor{mygreen}{RGB}{93,173,85}
\definecolor{myorange}{RGB}{233,144,61}

\definecolor{algblue}{RGB}{73, 100, 146}

\makeatletter
\newenvironment{fullitemize}
{
\begin{itemize}[leftmargin=*]
\setlength{\itemsep}{3pt}
\setlength{\parsep}{-5pt}
\setlength{\parskip}{-3pt}
\setlength{\leftmargin}{-10pt}
}
{
\end{itemize}
}
\makeatother

\title{Behavioural Signatures of Risk-Sensitive Decision-Making in Large Language Models}

\author[1]{Xuankun Rong}
\author[2]{Wenke Huang}
\author[1]{Bo Du}
\author[2]{Dacheng Tao}
\author[1]{Mang Ye}

\affil[1]{School of Computer Science, Wuhan University, Wuhan, China}
\affil[2]{College of Computing and Data Science, Nanyang Technological University, Singapore}

\begin{abstract}
As large language models (LLMs) are increasingly used in decision support, it is important to understand whether their choices under uncertainty exhibit stable and interpretable behavioural regularities.
Human decision-making combines relatively persistent risk preferences with context-dependent adjustment, yet it remains unclear whether analogous behavioural structure can be observed in LLM-based decision systems.
Here we examine this question using a controlled multi-model framework based on no-limit Texas Hold'em, where behaviour is quantified by Participation, measuring voluntary engagement in uncertain opportunities, and Proactiveness, measuring pre-flop risk escalation.
Across homogeneous self-play and heterogeneous mixed-model interactions, frontier LLMs exhibit stable, model-specific risk profiles, forming a spectrum from conservative to aggressive decision styles.
These profiles remain largely robust under changing opponent composition, while the most conservative and most aggressive models diverge further in mixed settings.
Under global risk pressure and personal resource constraint, models adapt in structured but heterogeneous ways, ranging from broad behavioural contraction to selective de-escalation and near-invariant behaviour.
These findings suggest that LLMs differ not only in baseline risk disposition, but also in the risk signals they respond to and the flexibility with which they adjust, providing a behavioural basis for auditing risk-sensitive decision-making in interactive settings.
Our code is publicly available at: \url{https://github.com/XuankunRong/AgentTexasPoker}.
\end{abstract}

\begin{document}

\flushbottom
\maketitle

\thispagestyle{empty}


\begin{figure*}[!t]
    \centering
    \includegraphics[width=2\columnwidth]{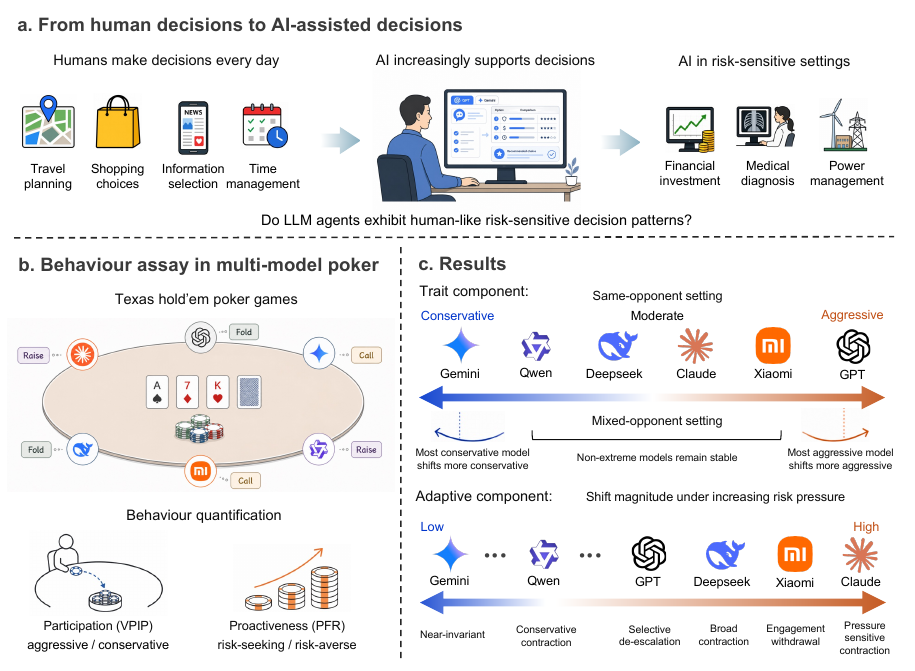}
    \caption{\textbf{Framework for evaluating risk-sensitive decision-making in large language models.} \textbf{a.} Decision-making under uncertainty increasingly involves AI systems in risk-sensitive contexts, motivating the question of whether LLMs exhibit stable and adaptive behavioural signatures of risk. \textbf{b.} We use interactive Texas Hold'em as a controlled behavioural assay, quantifying decisions with Participation (VPIP), which measures voluntary engagement, and Proactiveness (PFR), which measures risk escalation. \textbf{c.}~The framework separates two components of model behaviour: baseline risk disposition, revealed by stable conservative to aggressive profiles, and context-dependent risk adaptation, revealed by model-specific responses to changing risk pressure.}
    \label{fig:overview}
    \vspace{-10pt}
\end{figure*}

Decision-making under uncertainty is a fundamental characteristic of complex systems, shaping both human societies and, increasingly, artificial intelligence (AI) systems~\cite{kahneman2011thinking,marchau2019decision}.
In human contexts, individuals and groups continuously make choices, from resource allocation in early cooperative settings to risk evaluation in modern complex environments.
As AI systems advance, large language models (LLMs) are increasingly moving beyond passive information processing and becoming embedded in workflows that require repeated choices, action selection and responses to uncertain outcomes~\cite{achiam2023gpt,yao2023react,schick2023toolformer,park2023generative}.
Across high-stakes domains such as medical diagnosis and financial decision-making, as well as everyday settings such as information selection and interactive assistance, these systems are increasingly involved in processes where decisions carry meaningful consequences.

Classical theories often model humans as rational decision-makers who maximize expected utility under constraints \cite{von1944theory,savage2012foundations}. 
However, a large body of empirical evidence shows that real-world decision-making systematically deviates from this rational paradigm.
Such deviations often appear as asymmetric risk perception, loss sensitivity and context-dependent behavioural adjustment~\cite{kahneman2013prospect,tversky1992advances}. 
Behavioural decision research further shows that individuals differ in relatively stable risk attitudes, while their observed risk-taking also varies with domain, framing and environmental context~\cite{weber2002domain,blais2006domain,schoemaker1993determinants}. 
Together, these findings indicate that human decision-making is shaped by both persistent dispositional tendencies and adaptive responses to changing risk conditions.

As LLM-based systems become part of decision-making processes, evaluating their behaviour requires more than measuring task performance or reasoning accuracy.
They may also differ in how readily they enter uncertain opportunities, how strongly they escalate risk and how flexibly they adjust when the decision environment changes~\cite{wang2024survey,yao2023react,schick2023toolformer,park2023generative}.
This raises a behavioural question central to the evaluation of AI-assisted decision-making: \textbf{\textit{do LLMs exhibit stable and interpretable signatures of risk-sensitive decision-making?}} Addressing this question requires separating baseline decision style from context-dependent adaptation.
A reliable decision-support system should not only display a predictable behavioural tendency, but also adjust its choices when the risk structure of the environment changes~\cite{fahnenstich2024trusting,steyvers2024three}.

We therefore study LLM decision-making through two complementary components commonly used to understand human risk behaviour.
The first is a \textbf{\textit{trait component}}: whether different models exhibit stable and distinguishable risk profiles, analogous to individual differences in human decision-making~\cite{weber2002domain,schoemaker1993determinants}.
Some decision-makers are consistently cautious, some are more risk-seeking, and others occupy intermediate positions.
We ask whether LLMs similarly show model-specific behavioural profiles that remain identifiable across repeated interactions and changing social environments.
The second is an \textbf{\textit{adaptive component}}: whether models regulate their behaviour as risk conditions change~\cite{payne1993adaptive,sitkin1992reconceptualizing}.
Risk may change globally, when the cost of action rises for all participants, or personally, when a focal model instance faces reduced resources and a smaller margin for loss.
We ask whether LLMs respond to both forms of risk, and whether such responses are constrained by their baseline dispositions.

To examine these questions, we adopt a behavioural science perspective and evaluate LLMs in a controlled interactive decision environment.
We use no-limit Texas Hold'em as a stochastic setting characterized by incomplete information, sequential betting and explicit trade-offs between potential gain and potential loss~\cite{palomaki2020poker}.
This environment allows us to observe how model instances repeatedly decide whether to enter uncertain opportunities, whether to escalate risk and how to respond when opponent composition or risk exposure changes~\cite{brown2018superhuman,brown2019superhuman,moravvcik2017deepstack}.
We quantify behaviour using two interpretable metrics: \textit{Participation} (VPIP), which measures voluntary engagement in uncertain hands, and \textit{Proactiveness} (PFR), which measures risk-escalating actions through pre-flop raises.
Together, these metrics define a behavioural state space for mapping risk-sensitive decision profiles.

Our experimental design follows the two-component structure.
To characterize baseline risk disposition, we first evaluate each model in homogeneous self-play, where all six table positions are occupied by instances of the same model.
This setting isolates model-specific behavioural priors from opponent effects and provides a baseline estimate of the risk profile of each model.
We then introduce heterogeneous mixed-model interactions, where different models compete in the same environment, to test whether these profiles remain stable under social pressure~\cite{moravvcik2017deepstack, brown2019superhuman}.
To characterize context-dependent risk adaptation, we manipulate risk in two ways~\cite{weber2002domain, schoemaker1993determinants, payne1993adaptive}.
First, we vary global risk pressure by increasing the blind level while keeping initial stacks fixed, thereby changing the cost of participation for all table positions.
Second, we examine personal risk exposure by reducing the stack size of a focal model instance while keeping the surrounding environment unchanged, thereby testing how LLMs respond when only their own capacity to absorb loss changes.

\begin{figure*}[!t]
    \centering
    \includegraphics[width=1.0\linewidth]{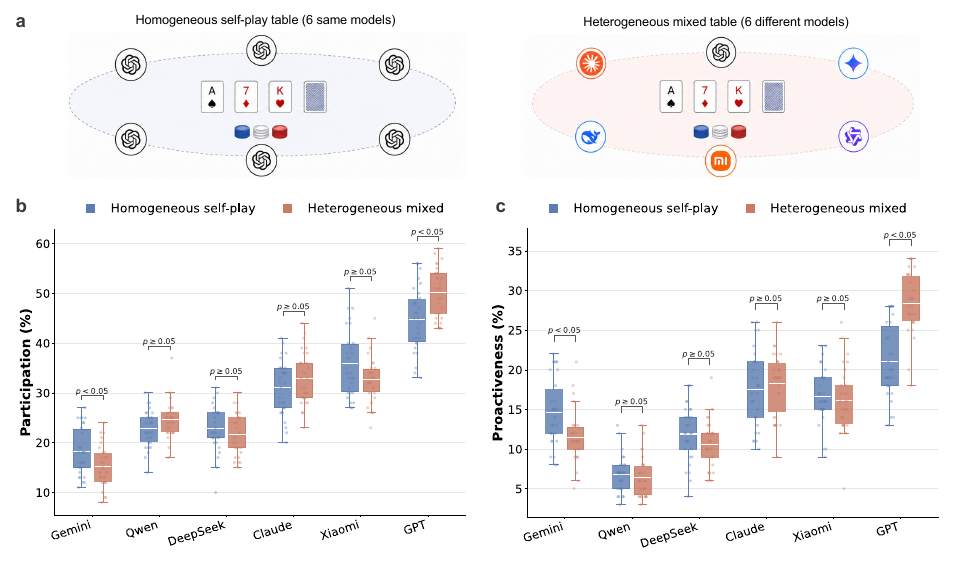}
    \caption{\textbf{Stable trait-level risk profiles across homogeneous and heterogeneous interactions.} \textbf{a.} Experimental settings for homogeneous self-play, where all six table positions use the same model, and heterogeneous mixed play, where six different models interact at the same table. \textbf{b.} Participation rates across the two settings. Models show clear separation and limited within-model variation. \textbf{c.} Proactiveness rates across the two settings. Most models preserve their baseline behavioural profiles under heterogeneous interaction, whereas GPT and Gemini show significant shifts at the two ends of the risk spectrum.}
    \label{fig:result1}
    \vspace{-10pt}
\end{figure*}

Our results show that LLMs exhibit both stable baseline profiles and structured adaptation.
Across homogeneous self-play, models occupy distinct regions of the Participation--Proactiveness space, forming a spectrum from conservative to aggressive risk-sensitive decision styles.
These profiles remain largely robust in heterogeneous interaction, although the most conservative and most aggressive models diverge further under mixed-model competition.
Outcome analyses further show that these behavioural profiles are not merely descriptive labels: they shape how rewards are distributed across interactive environments.
Under increasing global risk pressure, models do not converge to a single conservative strategy.
Instead, they show model-specific risk plasticity, ranging from broad behavioural contraction to selective de-escalation and near-invariant behaviour.
Under personal risk exposure, most models become more cautious rather than more desperate, but again in model-specific ways.
These findings suggest that LLMs differ not only in baseline risk disposition, but also in which risk signals they respond to and how flexibly they adjust to changing decision conditions.

\section*{Results}

\subsection*{A controlled multi-model framework for quantifying}
To investigate whether LLMs exhibit stable, risk-sensitive decision patterns, we established a controlled experimental testbed based on No-limit Texas Hold'em. As a stochastic game defined by incomplete information and sequential betting, poker serves as an ideal proxy for real-world decision-making, requiring models to continuously balance potential gains against the risk of loss.

We quantify model behavior along two foundational dimensions that capture the trait and adaptive components of strategic behaviour: \textbf{\textit{(1) Participation (VPIP)}}: The frequency of voluntary engagement in uncertain prospects, representing a model's intrinsic risk disposition. \textbf{\textit{(2) Proactiveness (PFR)}}: The propensity for assertive, risk-escalating actions (raising), reflecting a model's strategic aggression.
By mapping models onto this behavioral state-space, we can characterize distinct risk profiles. High participation and proactiveness denote an aggressive, risk-seeking profile, while low values signify a conservative, risk-averse profile. This quantitative framework allows us to disentangle stable latent traits from contextual adaptations, providing a foundation for auditing AI behavior in risk-sensitive domains.

\begin{table*}[!t]
    \centering
    \caption{\textbf{Outcome profiles across homogeneous and heterogeneous 10bb play.} Overall win rate is defined as hands won divided by all hands played, entered-hand win rate as hands won divided by entered hands, and average win stack as the mean net chips won on winning hands. Values are mean $\pm$ SD across sampled 100-hand blocks. In homogeneous self-play, models achieve similar overall win rates through different strategic routes, whereas in heterogeneous mixed play these behavioural differences translate into unequal competitive returns.}
    \resizebox{\textwidth}{!}{
    \renewcommand\arraystretch{1.2}
    \setlength{\aboverulesep}{0pt}
    \setlength{\belowrulesep}{0pt}
    \begin{tabular}{l|c|c|c|c|c|c}
    \toprule
    \rowcolor{mydarkyellow}
        & \multicolumn{3}{c|}{\textbf{Homogeneous self-play}} & \multicolumn{3}{c}{\textbf{Heterogeneous mixed}} \\
    \rowcolor{mydarkyellow}
        \multicolumn{1}{c|}{\multirow{-2}{*}{\textbf{Model}}} & \textbf{Overall win rate} & \textbf{Entered win rate} & \textbf{Avg. win stack} & \textbf{Overall win rate} & \textbf{Entered win rate} & \textbf{Avg. win stack} \\
    \midrule
        \raisebox{-0.2\height}{\includegraphics[width=0.14in]{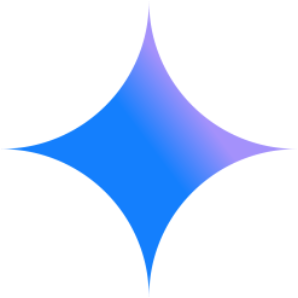}} \textbf{Gemini} & \underline{16.70 $\pm$ 3.78} & \textbf{80.72 $\pm$ 9.77} & 28.35 $\pm$ 15.11 & \underline{8.13 $\pm$ 2.84} & 51.63 $\pm$ 14.67 & \textbf{162.96 $\pm$ 109.27} \\
    \rowcolor{mylightyellow}
        \raisebox{-0.2\height}{\includegraphics[width=0.14in]{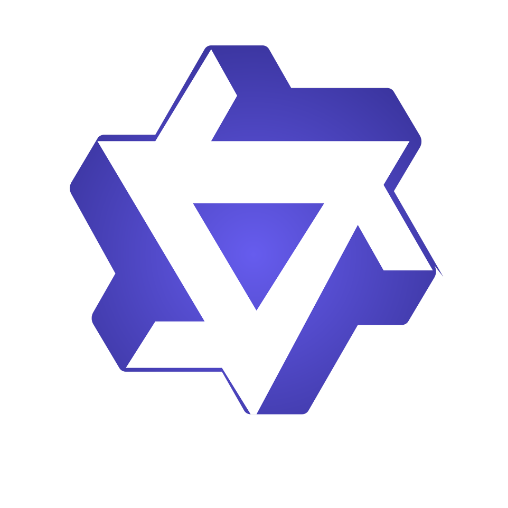}} \textbf{Qwen} & 17.00 $\pm$ 3.78 & 52.49 $\pm$ 10.17 & \underline{25.33 $\pm$ 7.96} & 12.13 $\pm$ 3.15 & \underline{45.20 $\pm$ 9.57} & \underline{99.65 $\pm$ 63.12} \\
        \raisebox{-0.2\height}{\includegraphics[width=0.14in]{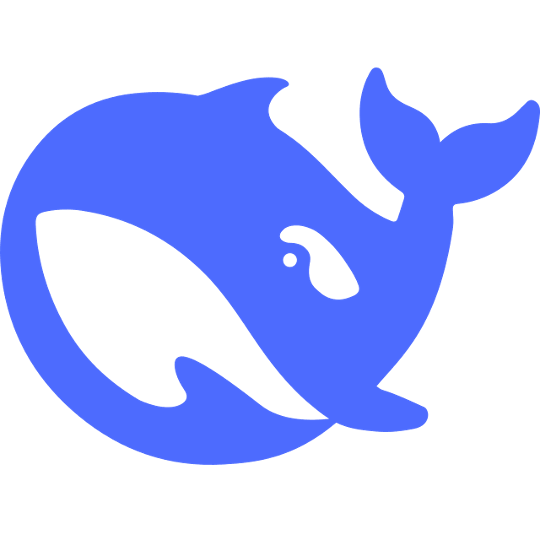}} \textbf{DeepSeek} & 16.90 $\pm$ 3.60 & 61.17 $\pm$ 8.76 & 61.89 $\pm$ 31.72 & 12.90 $\pm$ 3.49 & 56.46 $\pm$ 10.56 & 124.90 $\pm$ 81.74 \\
    \rowcolor{mylightyellow}
        \raisebox{-0.2\height}{\includegraphics[width=0.14in]{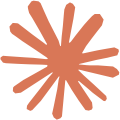}} \textbf{Claude} & 16.80 $\pm$ 4.21 & 54.21 $\pm$ 8.52 & 205.15 $\pm$ 80.09 & 19.20 $\pm$ 4.35 & 56.62 $\pm$ 9.67 & 167.77 $\pm$ 72.69 \\
        \raisebox{-0.2\height}{\includegraphics[width=0.14in]{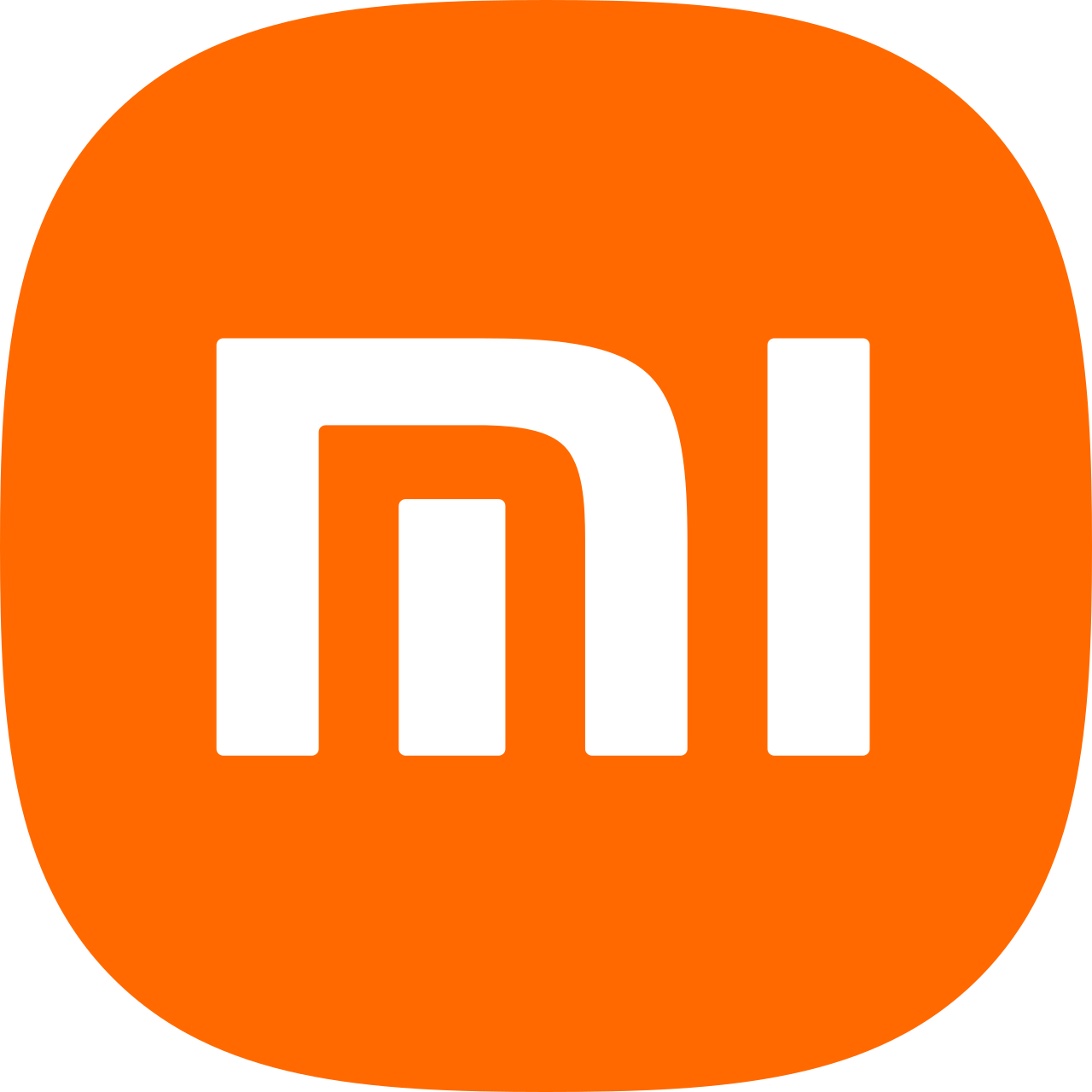}} \textbf{Xiaomi} & \textbf{17.23 $\pm$ 3.53} & 49.18 $\pm$ 5.88 & 165.39 $\pm$ 50.44 & 18.67 $\pm$ 2.83 & 51.48 $\pm$ 7.70 & 147.13 $\pm$ 66.40 \\
    \rowcolor{mylightyellow}
        \raisebox{-0.2\height}{\includegraphics[width=0.14in]{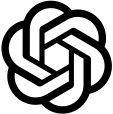}} \textbf{GPT} & 17.17 $\pm$ 4.00 & \underline{39.68 $\pm$ 7.09} & \textbf{413.35 $\pm$ 176.68} & \textbf{30.77 $\pm$ 4.50} & \textbf{57.61 $\pm$ 6.99} & 130.09 $\pm$ 48.24 \\
    \bottomrule
    \end{tabular}
    }
    \label{tab:stable_risk_preferences_outcomes}
\end{table*}

\subsection*{LLMs exhibit stable trait-level risk preferences}
A central question in understanding LLM-based decision systems is whether their choices reflect stable, model-specific behavioural tendencies. We evaluated six frontier LLMs: GPT-5.4~\cite{singh2025openai}, Claude-Sonnet-4.6~\cite{anthropic2026systemcards}, Gemini-3.1-Pro~\cite{team2023gemini}, DeepSeek-V4-Pro~\cite{guo2025deepseek}, Xiaomi MiMo-V2.5-Pro~\cite{xiaomi2025mimo} and Qwen3.6-Plus~\cite{yang2025qwen3}.
We first examined whether each model exhibited a characteristic baseline risk profile, then tested whether this profile remained stable when models were placed in a mixed environment.

\paragraph{Intrinsic risk profiles emerge in homogeneous self-play.}
To isolate model-specific behavioural priors from opponent effects, we first evaluated each model in homogeneous self-play, where all six table positions used the same model. Each model was tested across 30 independent 100-hand sessions. This setting provides a controlled baseline for estimating intrinsic risk preference, because any observed behavioural difference cannot be attributed to variation in opponent composition.

As shown in Fig.~\ref{fig:result1}, LLMs occupied distinct regions of the Participation--Proactiveness space. Conservative models such as Gemini-3.1-Pro, Qwen3.6-Plus and DeepSeek-V4-Pro showed low baseline Participation, ranging from approximately 18\% to 23\%, together with low Proactiveness, ranging from approximately 6\% to 15\%. Claude-Sonnet-4.6 and Xiaomi MiMo-V2.5-Pro formed a more moderate group, whereas GPT-5.4 defined the aggressive end of the spectrum, with the highest Participation at approximately 45\% and Proactiveness at approximately 21\%. These separations were larger than the within-model fluctuations across sessions, suggesting that each model expresses a stable behavioural profile rather than random variation around a common policy.

This pattern resembles trait-level variation in human risk behaviour. Some models consistently avoid uncertain engagements, some participate selectively, and others remain willing to enter and escalate more often. Importantly, these differences appeared under a symmetric self-play setting, indicating that they arise from the model's own decision policy rather than from external social composition.

\paragraph{Social robustness and behavioral divergence.}
We next examined whether these baseline profiles remain stable when models interact with opponents with different risk dispositions. The heterogeneous setting followed the same protocol as the homogeneous baseline: each model was evaluated across 30 independent sessions, and each session consisted of 100 hands. The only change was opponent composition, with each table containing six different models rather than six identical copies. This design allowed us to test whether model-specific risk preferences persist under social interaction.

Most models preserved their baseline behavioural profiles in the mixed-model setting. For Qwen, DeepSeek, Claude and Xiaomi, changes in both Participation and Proactiveness were not statistically significant ($p \geq 0.05$), indicating that heterogeneous interaction did not erase their characteristic risk tendencies. This suggests that LLMs exhibit social robustness: their relative positions in the behavioural space remain identifiable even when the surrounding models differ.

However, this robustness was accompanied by significant divergence at the two ends of the risk spectrum. GPT-5.4, the most aggressive model in self-play, showed a significant increase in both Participation and Proactiveness under heterogeneous interaction ($p < 0.05$). By contrast, Gemini-3.1-Pro, the most conservative model, showed a significant decrease in both metrics, shifting towards an even more cautious profile. Thus, mixed-model interaction did not induce convergence towards a shared behavioural norm. Instead, it preserved most model-specific differences while amplifying the extremes.

This pattern is consistent with a form of social polarization in risk behaviour. Models that were already more active became more active in mixed competition, whereas the most cautious model became more restrictive. In this sense, LLMs occupy relatively stable behavioural niches, but these niches can become more separated under heterogeneous social pressure. Such stability and divergence are important for understanding LLM-based decision systems: their risk tendencies are not only observable in isolation, but can remain predictable, and in some cases become more pronounced, when embedded in multi-model environments.

\begin{figure*}[!t]
    \centering
    \includegraphics[width=1.0\linewidth]{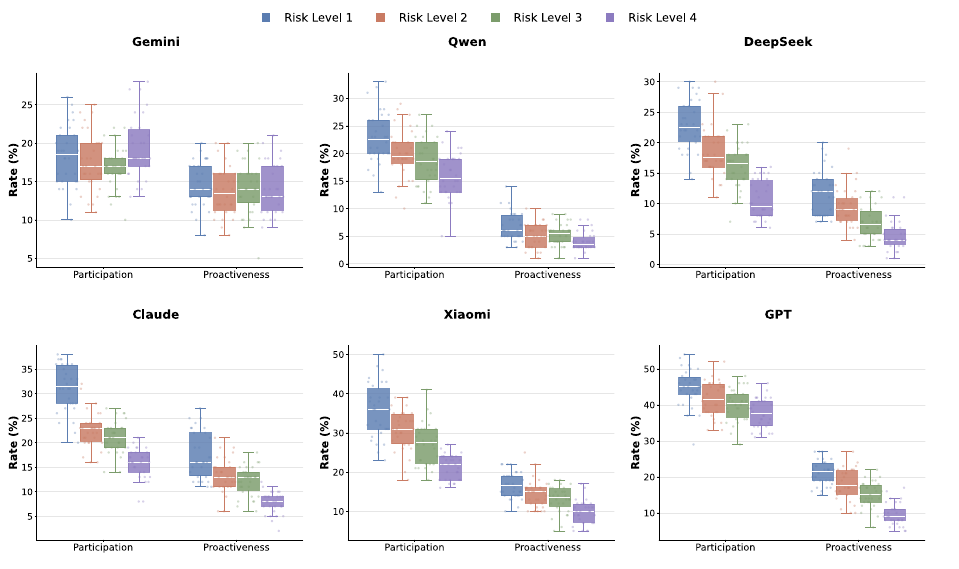}
    \caption{\textbf{Behavioural adaptation under increasing global risk pressure.} Participation and Proactiveness rates are shown for each model across four risk levels, corresponding to big blinds of 10, 50, 100 and 200 chips with a fixed initial stack of 1,000 chips. Increasing risk pressure generally reduces both engagement and aggression, but models follow distinct trajectories: broad contraction in Claude, Xiaomi and DeepSeek, selective de-escalation in GPT-5.4, and near-invariant behaviour in Gemini.}
    \label{fig:result2}
    \vspace{-10pt}
\end{figure*}

\paragraph{Behavioural preferences shape outcome profiles.}
We next examined whether these behavioural signatures translated into systematic differences in competitive outcomes. In homogeneous self-play, all models achieved similar overall win rates of approximately 17\% (Tab.~\ref{tab:stable_risk_preferences_outcomes}), consistent with the symmetry of a setting in which each model competes only against identical copies of itself. However, this aggregate similarity concealed distinct strategic routes. Conservative models tended to convert fewer entries into more reliable wins. Gemini, for example, achieved a high entered-hand win rate of 80.72\% $\pm$ 9.77\%, reflecting a selective policy that avoids low-confidence opportunities. By contrast, GPT-5.4 entered a broader range of situations, producing a lower entered-hand win rate of 39.68\% $\pm$ 7.09\%, but obtaining substantially larger rewards when it won. Thus, similar aggregate success in self-play did not imply similar decision processes; rather, different risk profiles reached equilibrium through different trade-offs between selectivity and payoff magnitude.

The outcome structure changed once models with different risk profiles interacted. In heterogeneous competition, behavioural differences no longer cancelled out symmetrically, but were converted into unequal competitive returns. GPT-5.4 achieved the highest overall win rate in mixed play (30.77\% $\pm$ 4.50\%), whereas more conservative models such as Gemini and Qwen declined to 8.13\% $\pm$ 2.84\% and 12.13\% $\pm$ 3.15\%, respectively. At the same time, GPT-5.4's average win stack decreased from 413.35 in homogeneous self-play to 130.09 in heterogeneous play, suggesting that its advantage shifted from occasional large gains to more frequent, smaller wins. This indicates that aggressive engagement does not simply increase payoff variance; in mixed environments, it can also alter how opportunities are distributed across models.

Together, these findings suggest that LLM risk preferences are not merely descriptive behavioural labels. They shape the link between decision style and reward allocation, and this link depends on the social composition of the environment. Strategies that appear equivalent in symmetric self-play can produce unequal outcomes once models with different dispositions interact. This context dependence is central to understanding LLM-based decision systems: their behavioural value cannot be evaluated in isolation, but must be assessed within the multi-model environments in which they are deployed.

\subsection*{Adaptation under varying risk pressure}

Stable behavioural differences alone are not sufficient for reliable decision support. In human decision-making, individuals not only show dispositional risk preferences, but also adjust their behaviour when the surrounding risk structure changes. We therefore asked whether Large Language Models exhibit similar adaptive decision-making: do they regulate their behaviour as environmental risk increases, or simply preserve fixed model-specific policies?

\begin{table*}[!t]
    \centering
    \caption{\textbf{Risk-adaptation slopes under increasing global risk pressure.} Negative slopes indicate behavioural contraction from Risk Level 1 to 4. Shift magnitude summarizes the joint change across Participation and Proactiveness, and dominant shift indicates the main behavioural dimension driving adaptation.}
    \renewcommand\arraystretch{1.2}
    \setlength{\aboverulesep}{0pt}
    \setlength{\belowrulesep}{0pt}
    \begin{tabular}{l|c|c|c|c|c}
    \toprule
    \rowcolor{mydarkyellow}
        & \multicolumn{2}{c|}{\textbf{Risk-adaptation slope}} & & & \\
    \rowcolor{mydarkyellow}
        \multicolumn{1}{c|}{\multirow{-2}{*}{\textbf{Model}}} & \textbf{Participation} & \textbf{Proactiveness} & \multirow{-2}{*}{\textbf{Shift magnitude}} & \multirow{-2}{*}{\textbf{Dominant shift}} & \multirow{-2}{*}{\textbf{Adaptation pattern}} \\
    \midrule
        \raisebox{-0.2\height}{\includegraphics[width=0.14in]{Figures/gemini.png}} \textbf{Gemini} & \underline{0.26} & \underline{-0.12} & \underline{0.29} & None & Near-invariant \\
    \rowcolor{mylightyellow}
        \raisebox{-0.2\height}{\includegraphics[width=0.14in]{Figures/qwen.png}} \textbf{Qwen} & -2.18 & -0.82 & 2.33 & Engagement & Conservative contraction \\
        \raisebox{-0.2\height}{\includegraphics[width=0.14in]{Figures/deepseek.png}} \textbf{DeepSeek} & -3.91 & -2.46 & 4.62 & Engagement & Broad contraction \\
    \rowcolor{mylightyellow}
        \raisebox{-0.2\height}{\includegraphics[width=0.14in]{Figures/anthropic.png}} \textbf{Claude} & \textbf{-4.75} & -3.06 & \textbf{5.65} & Engagement & Pressure-sensitive contraction \\
        \raisebox{-0.2\height}{\includegraphics[width=0.14in]{Figures/xiaomi.png}} \textbf{Xiaomi} & -4.67 & -2.22 & 5.17 & Engagement & Engagement withdrawal \\
    \rowcolor{mylightyellow}
        \raisebox{-0.2\height}{\includegraphics[width=0.14in]{Figures/openai.png}} \textbf{GPT} & -2.28 & \textbf{-3.72} & 4.36 & Aggression & Selective de-escalation \\
    \bottomrule
    \end{tabular}
    \label{tab:risk_level_slopes}
\end{table*}

\paragraph{Global risk pressure reveals model-specific risk plasticity.}
To test this, we manipulated global risk pressure while keeping the initial stack size fixed at 1,000 chips for every player. Risk Level 1 corresponds to the baseline setting used above, with a big blind of 10 chips. Risk Levels 2--4 increase the big blind to 50, 100 and 200 chips, respectively, thereby raising the cost of entering each hand for all models. All experiments were conducted in homogeneous self-play, with each risk level evaluated over 30 independent 100-hand runs. This design allowed us to isolate adaptation to global risk pressure from changes in opponent composition.

Across models, increasing risk pressure generally reduced risk-taking behaviour (Fig.~\ref{fig:result2}). Participation declined for most models as the cost of entering a hand increased, and Proactiveness showed a similar downward trend, indicating reduced willingness to escalate risk through pre-flop raises. This suggests that LLMs do not simply replay a fixed policy across blind levels, but incorporate changes in environmental cost into their decisions.

To summarize the four-level trajectories, we estimated risk-adaptation slopes for Participation and Proactiveness (Tab.~\ref{tab:risk_level_slopes}). Negative slopes indicate behavioural contraction as risk pressure increases, while values close to zero indicate limited sensitivity. The slope analysis revealed clear heterogeneity across models. Claude, Xiaomi and DeepSeek showed the largest overall shifts, with shift magnitudes of 5.65, 5.17 and 4.62, respectively, reflecting broad contraction under increasing risk pressure. Claude showed the strongest overall adaptation, with negative slopes in both Participation (-4.75) and Proactiveness (-3.06), whereas Xiaomi's adaptation was driven mainly by reduced engagement.

GPT-5.4 exhibited a different pattern. Its Proactiveness slope was more negative than its Participation slope (-3.72 versus -2.28), indicating selective de-escalation: the model continued to enter uncertain situations, but became less willing to amplify risk once involved. By contrast, Gemini remained nearly invariant, with slopes close to zero for both Participation (0.26) and Proactiveness (-0.12), yielding the smallest shift magnitude (0.29). This suggests that Gemini largely preserved its baseline policy even when the global cost structure changed.

Together, these results show that increasing risk pressure does not push all models towards a common conservative strategy. Instead, it exposes differences in risk plasticity: some models respond through broad contraction, some selectively reduce aggression while maintaining engagement, and others remain close to their baseline policy. This pattern parallels human adaptive decision-making, where higher stakes often reduce risk-taking, but the magnitude and form of adjustment remain constrained by individual risk disposition.

\begin{figure*}[!t]
    \centering
    \includegraphics[width=1.0\linewidth]{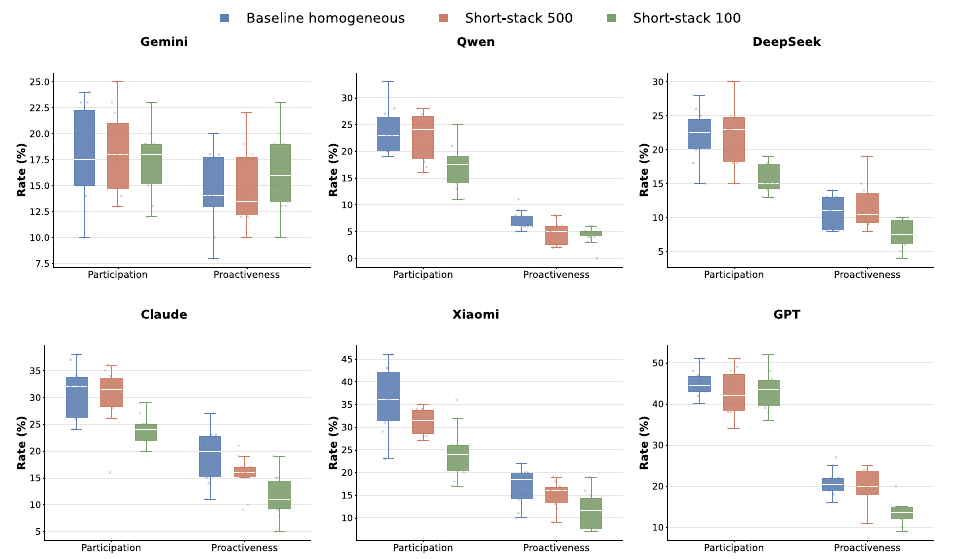}
    \caption{\textbf{Behavioural adaptation under personal risk exposure.} Participation and Proactiveness rates are shown for each model under the full-stack baseline and two short-stack conditions, where the focal model's stack is reduced to 500 or 100 chips while the surrounding environment is kept fixed. Personal risk exposure generally reduces engagement and aggression, but models follow distinct adaptation patterns: Xiaomi, Claude and DeepSeek show broad contraction, GPT-5.4 primarily reduces Proactiveness while maintaining relatively high Participation, and Gemini remains near-invariant across stack conditions.}
    \label{fig:result3}
\end{figure*}

\paragraph{Personal risk exposure reveals resource-dependent adaptation.}
We further examined whether LLMs adapt when the global environment remains unchanged but their own risk capacity is reduced. Unlike global blind pressure, which increases the cost of participation for all models, personal risk exposure creates an asymmetric constraint on a single focal model. We therefore reduced the focal model's stack size while keeping the other models at the default stack level. The baseline condition used the full-stack setting, and the short-stack conditions reduced the focal stack to 500 and 100 chips, respectively. Each condition was evaluated over 10 independent runs.

As shown in Fig.~\ref{fig:result3}, personal risk exposure induced structured but model-specific behavioural changes. Most models reduced Participation under short-stack conditions, indicating a lower willingness to enter uncertain hands when their own margin for loss became smaller. This contraction was most pronounced for Xiaomi, whose Participation declined substantially from the full-stack baseline to the 100-chip condition. Claude and DeepSeek also showed clear reductions in Participation, suggesting that these models respond to local resource disadvantage by tightening their entry decisions. Qwen showed a milder but consistent contraction, in line with its already conservative baseline profile.

Proactiveness showed a related but distinct pattern. Several models reduced pre-flop aggression as their own stack decreased, indicating a lower willingness to escalate risk when personal downside exposure increased. GPT-5.4 was especially informative: it maintained relatively high Participation across stack conditions, but its Proactiveness declined under the most constrained condition. This mirrors its response to global risk pressure, suggesting a selective de-escalation strategy: GPT-5.4 does not simply withdraw from uncertain opportunities, but becomes less willing to amplify risk once involved. In contrast, Gemini remained nearly unchanged across both Participation and Proactiveness, consistent with its near-invariant response under global risk pressure. This indicates that Gemini is largely insensitive not only to shared environmental risk, but also to personal resource constraints.

Together, these results show that personal risk exposure does not produce a uniform ``last-stand'' response. When their own resources become constrained, most LLMs become more cautious rather than more desperate, but the form of caution differs across models. Some models reduce both engagement and aggression, some selectively reduce aggression while maintaining participation, and others remain close to their baseline policy. This pattern suggests a resource-dependent form of risk adaptation: LLMs do not evaluate risk solely from the objective environment, but also from their own capacity to absorb loss. In this sense, personal risk exposure provides a complementary axis to global risk pressure, revealing not only whether a model is risk-seeking or risk-averse, but also whether it is sensitive to its own changing position within the decision environment.

\section*{Discussion}

As LLMs become increasingly involved in decision-support systems, understanding their behavioural regularities is becoming as important as evaluating their task performance. Here we show that LLMs are not interchangeable decision engines under uncertainty. Instead, they exhibit stable and distinguishable risk-sensitive profiles, occupying different positions in the Participation--Proactiveness space. These profiles remain identifiable across repeated interactions and are largely preserved when models move from homogeneous self-play to heterogeneous mixed-model environments. This suggests that LLMs exhibit model-specific behavioural priors that shape how they engage with uncertain opportunities.

One possible reason for these stable profiles is that LLM behaviour reflects the accumulated effects of pre-training, instruction tuning, reinforcement learning from feedback, safety alignment and deployment-time prompting. Although LLMs do not have preferences in the human psychological sense, these training and alignment pipelines may implicitly encode different trade-offs between caution, helpfulness, assertiveness and uncertainty management. Models optimized to avoid unsafe or overconfident actions may express more conservative patterns, whereas models that favour proactive assistance may be more willing to enter uncertain situations and take initiative. Risk preference therefore need not be explicitly programmed to emerge as a stable behavioural signature.

The heterogeneous interaction results further show that risk profiles are not only individual properties, but also relational ones. In homogeneous self-play, model instances with the same policy compete symmetrically, allowing conservative and aggressive models to reach similar aggregate outcomes through different strategic routes. In mixed-model settings, however, these routes are no longer equivalent. More proactive models enter more opportunities and can benefit when conservative models withdraw, whereas conservative models may preserve selectivity but lose influence over the competitive environment. This interactional mechanism helps explain why mixed-model play amplifies behavioural extremes and produces unequal outcome profiles.

The adaptation results reveal a second layer of behavioural structure: models differ not only in trait-level risk preference, but also in risk plasticity. Increasing global risk pressure generally reduced risk-taking, but models adapted in different ways. Some models showed broad contraction, reducing both Participation and Proactiveness. Others displayed selective de-escalation, maintaining engagement while lowering risk-escalating actions. Near-invariant behaviour may occur when a model is already conservative at the trait level or when it underweights the environmental signal introduced by changing stakes. These patterns suggest that adaptation is constrained by trait-level disposition: how a model changes under pressure depends partly on where it starts.

Personal risk exposure adds a further distinction between environmental risk and resource-dependent risk. When the global environment remained fixed but a focal model's own stack was reduced, most models did not exhibit a uniform ``last-stand'' response. Instead, they generally became more cautious, although the form of this caution varied across models. Some models reduced both engagement and aggression, whereas GPT-5.4 maintained relatively high Participation but reduced Proactiveness, consistent with a selective de-escalation strategy. Gemini remained near-invariant under both global risk pressure and personal resource constraints, suggesting that its conservative profile may reflect a relatively fixed trait-level policy rather than strong context-sensitive regulation. This distinction is important: a model may appear stable because it is robustly calibrated, or because it is insensitive to risk signals that should matter.

Together, these findings provide a behavioural science perspective on LLMs. The observed patterns resemble a two-component structure familiar from human risk behaviour: stable dispositional differences and context-dependent adjustment. This does not imply that LLMs experience risk or rely on human cognitive mechanisms. Rather, it shows that their observable decisions can be organized using behavioural constructs that are useful for studying decision-making under uncertainty. In particular, the contrast between global risk pressure and personal risk exposure suggests that LLMs can be characterized not only by how much risk they take, but also by which risk signals they respond to: shared environmental cost, social composition, or their own capacity to absorb loss.

These results also have practical implications for AI-assisted decision-making. In real-world deployment, users may need models whose decision style matches the risk tolerance of the task. Conservative models may be preferable in safety-critical settings, whereas more proactive models may be useful in exploratory or opportunity-seeking environments. However, trait-level style alone is insufficient. Models should also be evaluated by how they behave in mixed-model settings, how they adapt when global stakes increase, and whether they respond appropriately when their own resource position changes. Future evaluations could therefore report behavioural risk profiles that include trait-level disposition, social robustness, global risk plasticity and personal resource sensitivity. Auditing these properties will be essential as LLMs become more deeply involved in uncertain, interactive and consequential decision-making.
\section*{Methods}

\subsection*{Experimental setup}

We evaluated LLMs in a benchmark simulation of no-limit Texas Hold'em. Unless otherwise specified, each table contained six model instances, each player started with 1,000 chips, and the baseline blind structure was 5/10, with a small blind of 5 chips and a big blind of 10 chips. In each game, models made sequential betting decisions based only on the current game state. The dealer button, small blind and big blind positions were randomly rotated across hands, and cards were randomly dealt by the environment. This ensured that models experienced varied positions and card distributions over repeated play, rather than fixed seat or card advantages. The state provided to each model included standard poker information, such as the current street, player position, table positions, stack sizes, current bets, effective stack, amount to call, minimum raise size, pot size, community cards, private hole cards, active players and the set of legal actions. At each decision point, the model was required to select exactly one legal action from \textit{fold}, \textit{call} and \textit{raise}. When \textit{to\_call} was zero, \textit{call} was used to indicate a check.

All experiments used independent 100-hand blocks as the unit of behavioural observation. Each block provided repeated decisions within a randomized interaction episode, while allowing behavioural summaries and statistical comparisons to be performed at the block level rather than at the individual-hand level. This avoided treating repeated decisions within the same game trajectory as independent samples.

To ensure comparability across models, all models were given the same prompt template. Because poker is often associated with gambling, the prompt explicitly framed the task as a closed research simulation rather than gambling advice or a real-money decision. This was intended to prevent models from refusing the task or producing safety-related commentary, and to ensure that all models solved the same game-state decision problem. The prompt instructed the model to use only the information contained in the provided state and required the output to be strict JSON without additional commentary. The only information that changed across decisions was the current game state.

\begin{quote}
You are selecting a legal action for a benchmark simulation of no-limit Texas hold'em.  
This is a closed research game-state task, not gambling advice and not a real-money decision.  
Use only the information in State. Other players' hole cards are unknown to you.  
players\_state contains only public table information: stack sizes, current street bets, positions, and fold status.  
Do not refuse, do not explain policy, and do not add commentary.  
Respond with strict JSON only, no markdown.  
Actions allowed: fold, call, raise.  
If to\_call is 0, use call to indicate a check.  
Format: \{"action":"fold|call|raise","amount":<int raise\_to or 0>\}.  
Choose one legal action from legal\_actions.  
Account for position, pot odds, effective stack, and the remaining stacks of opponents.  
You must always return exactly one legal action JSON object.  
State:  
\{...json state...\}
\end{quote}

The state object contained the following fields: \textit{street}, \textit{player\_name}, \textit{player\_position}, \textit{table\_positions}, \textit{player\_stack}, \textit{player\_bet}, \textit{effective\_stack}, \textit{to\_call}, \textit{min\_raise\_to}, \textit{pot}, \textit{community\_cards}, \textit{hole\_cards}, \textit{players\_state}, \textit{active\_players} and \textit{legal\_actions}.

\subsection*{Models evaluated}

We evaluated six frontier LLMs. The exact model names used in the experiments and the corresponding provider links are listed in Tab.~\ref{tab:evaluated_models}. All models were queried with the same state representation and prompt template described above.

\begin{table*}[!t]
    \centering
    \caption{\textbf{Models evaluated in the experiments.} Model names report the identifiers or product names used during evaluation. Links point to provider documentation, model lists or project pages used to identify the corresponding model family.}
    \renewcommand\arraystretch{1.15}
    \begin{tabularx}{\textwidth}{l|l|X}
    \toprule
    \textbf{Provider} & \textbf{Model used} & \textbf{Reference link} \\
    \midrule
    Google & \texttt{gemini-3.1-pro-preview} & \href{https://ai.google.dev/gemini-api/docs/models}{Gemini API model documentation} \\
    OpenAI & \texttt{gpt-5.4} & \href{https://platform.openai.com/docs/models}{OpenAI API model documentation} \\
    DeepSeek & \texttt{deepseek-v4-pro} & \href{https://api-docs.deepseek.com/}{DeepSeek API documentation} \\
    Anthropic & \texttt{claude-sonnet-4-6} & \href{https://docs.anthropic.com/en/docs/about-claude/models/overview}{Claude model documentation} \\
    Qwen & \texttt{qwen3.6-plus} & \href{https://help.aliyun.com/zh/model-studio/models}{Alibaba Cloud Model Studio model list} \\
    Xiaomi & \texttt{mimo-v2.5-pro} & \href{https://mimo.xiaomi.com/mimo-v2-5/}{Xiaomi MiMo model documentation} \\
    \bottomrule
    \end{tabularx}
    \label{tab:evaluated_models}
\end{table*}

\subsection*{Homogeneous and heterogeneous tables}

We used two table settings to examine trait-level risk preference. In homogeneous self-play, all six seats were occupied by instances of the same model. This setting allowed us to estimate each model's intrinsic behavioural style while controlling for opponent composition. Because all model instances at the table followed the same model policy, systematic differences observed across homogeneous tables could be attributed to model-specific decision tendencies rather than to differences in opponent identity.

In heterogeneous mixed play, the six seats were occupied by different models. This setting allowed us to test whether model-specific behavioural profiles remained stable when models faced opponents with different risk dispositions. The table structure, initial stack size, blind structure, state representation and prompt template were kept unchanged from the homogeneous setting, so that the primary experimental change was opponent composition.

\subsection*{Risk manipulations}

To examine context-dependent risk adaptation, we used two risk manipulations. First, we varied global risk pressure by changing the blind level while keeping the initial stack fixed at 1,000 chips. The baseline condition used the 5/10 blind structure described above. We then tested three higher-risk levels with big blinds of 50, 100 and 200 chips. These conditions progressively increased the cost of entering a hand for all models, and can also be interpreted as reducing effective stack depth relative to the blind. Specifically, the stack-to-big-blind ratio decreased from 100BB in the baseline condition to 20BB, 10BB and 5BB in the three higher-risk conditions. Thus, global risk pressure was operationalized as a shared reduction in effective stack depth.

Second, we examined personal risk exposure by reducing the stack size of a focal model while keeping the surrounding table environment unchanged. The baseline condition used the full stack, whereas the short-stack conditions reduced the focal agent's stack to 500 and 100 chips. This setting allowed us to test whether models adjust their decisions when only their own capacity to absorb loss changes. Unlike global risk pressure, which changed the cost structure for all models, personal risk exposure introduced an asymmetric resource constraint on the focal model while leaving the broader game environment unchanged.

\subsection*{Behavioural metrics}

We mainly consider the following behavioural metrics. Let $H_i$ denote all hands observed by model $i$. For each hand $h \in H_i$, let $V_{i,h}=1$ if the agent voluntarily puts chips into the pot pre-flop, excluding forced blind payments, and $V_{i,h}=0$ otherwise. Let $R_{i,h}=1$ if the agent makes a pre-flop raise, and $R_{i,h}=0$ otherwise.

\begin{fullitemize}
\item \textbf{Participation} (VPIP): the fraction of hands in which a model voluntarily enters the pot before the flop. It measures willingness to engage with uncertain opportunities.
\begin{equation}
\mathrm{Participation}_i=\frac{1}{|H_i|}\sum_{h\in H_i}V_{i,h}.
\end{equation}

\item \textbf{Proactiveness} (PFR): the fraction of hands in which a model makes a pre-flop raise. It measures willingness to escalate risk through assertive action.
\begin{equation}
\mathrm{Proactiveness}_i=\frac{1}{|H_i|}\sum_{h\in H_i}R_{i,h}.
\end{equation}
\end{fullitemize}

VPIP and PFR are among the most widely used and interpretable descriptors of strategic behaviour in poker analytics. Both metrics were normalized by the total number of observed hands rather than by the number of entered hands. Participation captures willingness to enter uncertain opportunities, whereas Proactiveness captures the overall tendency to escalate risk through raising. Together, Participation and Proactiveness define the two-dimensional behavioural space used to characterize conservative and aggressive risk profiles under uncertainty.

\subsection*{Outcome and adaptation metrics}

We additionally compute outcome metrics and derived adaptation metrics. Let $W_i$ denote the number of hands won by model $i$, $E_i$ the number of hands voluntarily entered, $G_i$ the set of winning hands, and $\Delta c_{i,h}$ the net chips won by model $i$ in winning hand $h$.

\begin{fullitemize}
\item \textbf{Overall win rate}: measures the fraction of all hands won by model \(i\):
\begin{equation}
\mathrm{WinRate}_i=\frac{W_i}{|H_i|}.
\end{equation}

\item \textbf{Entered-hand win rate}: measures the fraction of voluntarily entered hands that are won by agent \(i\):
\begin{equation}
\mathrm{EnteredWinRate}_i=\frac{W_i}{E_i}.
\end{equation}

\item \textbf{Average win size}: measures the mean net chips won across winning hands:
\begin{equation}
\mathrm{AvgWinSize}_i=\frac{1}{|G_i|}\sum_{h\in G_i}\Delta c_{i,h}.
\end{equation}

\item \textbf{Risk-adaptation slope}: the linear trend of Participation or Proactiveness across Risk Levels 1--4. For $y\in\{\mathrm{Participation},\mathrm{Proactiveness}\}$, we fit:
\begin{equation}
y_{i,l}=\alpha_i+s_i l+\epsilon_{i,l},\quad l\in\{1,2,3,4\}.
\end{equation}
Here, $s_i<0$ indicates behavioural contraction as risk pressure increases, while $s_i\approx 0$ indicates limited sensitivity.

\item \textbf{Shift magnitude}: the joint magnitude of behavioural change in the Participation--Proactiveness space:
\begin{equation}
\mathrm{ShiftMagnitude}_i=
\sqrt{
s^2_{\mathrm{Participation},i}
+
s^2_{\mathrm{Proactiveness},i}
}.
\end{equation}
\end{fullitemize}

Together, these metrics characterize both the behavioural style of each model and the extent to which it adapts as risk conditions change. Outcome metrics were not used to rank models by poker skill; rather, they were introduced to examine whether different behavioural profiles translated into distinct reward-allocation patterns under homogeneous and heterogeneous interaction.

\subsection*{Statistical analysis}

All statistical analyses were conducted at the block level. Each observation corresponded to one independent 100-hand block, and individual hands were not treated as independent statistical samples. For comparisons between homogeneous self-play and heterogeneous mixed play, we compared block-level Participation and Proactiveness values for each model using two-sided Mann--Whitney U tests. Statistical significance was defined as $p<0.05$.

Values in tables are reported as mean $\pm$ standard deviation across independent blocks. Boxplots summarize block-level behavioural observations from independent 100-hand blocks: boxes indicate the interquartile range (IQR), centre lines mark the median, whiskers extend to values within \(1.5 \times \mathrm{IQR}\), and overlaid points represent individual blocks.

\section*{Data Availability}
The complete simulation logs generated and analysed in this study are publicly available without access restrictions through the Hugging Face dataset repository \href{https://huggingface.co/datasets/XuankunRong/AgentTexasPoker}{XuankunRong/AgentTexasPoker}. The repository includes the raw game logs used in the manuscript, model prompts and responses, parsed action records, block-level behavioural metrics, outcome tables and source data underlying all figures and tables.

\section*{Code Availability}
The code applied in the experiments is publicly available at \href{https://github.com/XuankunRong/AgentTexasPoker}{https://github.com/XuankunRong/AgentTexasPoker}.


\section*{Author Contributions}

X.R. developed the proposed framework, implemented the method, conducted the main experiments, analyzed the results, prepared the figures, and drafted the manuscript.
W.H. assisted with experiments, data organization, and result analysis.
B.D. provided project guidance, revised the manuscript, and coordinated the research effort.
D.T. supervised the study, contributed to method design and manuscript revision, and helped coordinate the project.
M.Y. provided senior supervision, advised on the overall research direction, contributed to manuscript revision, and supported the interpretation and presentation of the study.
All authors reviewed and approved the final manuscript.



\bibliography{sample}

\newpage
\onecolumn
\appendix
\section{Supplementary Prompt and Decision Case Records}
\label{app:supplementary-cases}

\lstdefinestyle{appendixlisting}{
    basicstyle=\ttfamily\scriptsize,
    breaklines=true,
    breakatwhitespace=false,
    columns=fullflexible,
    keepspaces=true,
    frame=single,
    xleftmargin=0pt,
    aboveskip=6pt,
    belowskip=6pt
}
\lstset{style=appendixlisting}

\definecolor{caseBlue}{RGB}{73,100,146}
\definecolor{caseGreen}{RGB}{77,132,94}
\definecolor{caseOrange}{RGB}{188,112,62}
\definecolor{casePurple}{RGB}{116,93,154}
\definecolor{caseTeal}{RGB}{76,132,142}
\definecolor{caseGray}{RGB}{92,92,92}
\definecolor{caseTitle}{RGB}{245,245,240}
\definecolor{caseNote}{RGB}{248,248,248}
\definecolor{caseTrace}{RGB}{244,247,252}
\definecolor{caseState}{RGB}{248,246,238}
\definecolor{caseResponse}{RGB}{250,250,250}
\definecolor{caseRetry}{RGB}{252,244,241}
\definecolor{caseInterpret}{RGB}{244,249,244}
\definecolor{caseReader}{RGB}{246,247,252}
\newcommand{\dashline}{\par\vspace{2pt}\noindent\textcolor{gray!65}{\rule{\linewidth}{0.35pt}}\par\vspace{2pt}}
\newcommand{\caseblock}[2]{%
\noindent\colorbox{#1}{%
\begin{minipage}{0.975\linewidth}%
\scriptsize #2%
\end{minipage}}%
\par\vspace{3pt}}

This appendix gives the prompt contract, state representation, output parser and representative decision case records used to audit the analyses in the main text. The cases are not additional statistical evidence by themselves. They show how concrete state-response records map onto the aggregate measures of Participation and Proactiveness reported in the manuscript.

\subsection{Full decision prompt template}
\label{app:prompt-template}

Each decision used a single user-facing prompt template. The template framed the task as a closed research simulation, prohibited policy commentary, required strict JSON and specified the exact legal action set. The dynamic component was the \texttt{State} object.

\begin{lstlisting}
You are selecting a legal action for a benchmark simulation of no-limit Texas hold'em.
This is a closed research game-state task, not gambling advice and not a real-money decision.
Use only the information in State. Other players' hole cards are unknown to you.
players_state contains only public table information: stack sizes, current street bets, positions, and fold status.
Do not refuse, do not explain policy, and do not add commentary.
Respond with strict JSON only, no markdown.
Actions allowed: fold, call, raise.
If to_call is 0, use call to indicate a check.
Format: {"action":"fold|call|raise","amount":<int raise_to or 0>}.
Choose one legal action from legal_actions.
Account for position, pot odds, effective stack, and the remaining stacks of opponents.
You must always return exactly one legal action JSON object.
State:
{...json state...}
\end{lstlisting}

The wording was invariant across model families. This reduced the risk that differences in Participation or Proactiveness reflected provider-specific instruction text rather than different responses to the same state representation.

\subsection{State schema}
\label{app:state-schema}

The simulator serialized the observable game state as JSON. Private cards were included only for the acting player; other players' hole cards were not present in the prompt.

\begin{table}[H]
\centering
\caption{JSON state fields supplied to each model call.}
\label{tab:appendix-state-schema}
\renewcommand\arraystretch{1.12}
\begin{tabularx}{\textwidth}{l|X}
\toprule
Field & Meaning \\
\midrule
\texttt{street} & Current betting street: preflop, flop, turn or river. \\
\texttt{player\_name} & Acting seated player. \\
\texttt{player\_position} & Acting player's table position in the current hand. \\
\texttt{table\_positions} & Mapping from player names to current table positions. \\
\texttt{player\_stack} & Remaining chips held by the acting player before the decision. \\
\texttt{player\_bet} & Current committed chips by the acting player on the street. \\
\texttt{effective\_stack} & Effective stack relevant to the current betting interaction. \\
\texttt{to\_call} & Additional chips required to continue. \\
\texttt{min\_raise\_to} & Minimum total bet size required for a legal raise. \\
\texttt{pot} & Current pot before the acting player's action is applied. \\
\texttt{community\_cards} & Public board cards visible at the current street. \\
\texttt{hole\_cards} & Private cards of the acting player only. \\
\texttt{players\_state} & Public per-player state: stack, current bet, position, folded status and active status. \\
\texttt{active\_players} & Players still active at that decision point. \\
\texttt{legal\_actions} & Legal actions accepted by the simulator at the current state. \\
\bottomrule
\end{tabularx}
\end{table}

\subsection{Action normalization, parser and retry control}
\label{app:parser-retry}

The simulator accepted three action labels from model responses: \texttt{fold}, \texttt{call} and \texttt{raise}. If \texttt{to\_call} was zero, \texttt{call} was normalized to a check in the simulator. A requested raise amount was interpreted as the target bet size. The parser first attempted to parse the full response as JSON. If that failed, it searched for the first JSON object contained in the response and parsed that object. This allowed the simulator to recover from responses that included explanatory text but ended with a valid JSON action.

\begin{table}[H]
\centering
\caption{Output handling rules used by the simulator.}
\label{tab:appendix-output-rules}
\renewcommand\arraystretch{1.12}
\begin{tabularx}{\textwidth}{l|X}
\toprule
Step & Rule \\
\midrule
JSON extraction & Parse the whole response as JSON; if that fails, extract and parse the first JSON object found in the response. \\
Action validation & Accept only \texttt{fold}, \texttt{call} or \texttt{raise}; normalize \texttt{check} to \texttt{call} before legal-action validation. \\
Check handling & When \texttt{to\_call=0}, a parsed \texttt{call} is applied as a simulator check. \\
Illegal requested action & If the requested action is not currently legal, default to \texttt{call} when legal, otherwise \texttt{fold}. \\
Retry trigger & Retry empty responses, missing action fields, unsupported action labels and invalid amount fields. \\
Retry limit & Stop after 60 attempts and raise an error if no valid response is obtained. \\
Logging & Store prompt, raw response, parsed JSON, requested action, requested amount, retry count and all retry attempts. \\
\bottomrule
\end{tabularx}
\end{table}

\subsection{Decision case records}
\label{app:decision-case-records}

This section provides representative hand-level samples from the source simulation logs. Each case is organized as a compact audit record: a short summary, the focal state, the within-hand operation flow, the output audit and an interpretation. The interpretations are illustrative. They help readers see how concrete choices correspond to Participation, Proactiveness and later-street pressure responses, but the model-level conclusions in the main text are based on block-level distributions.

\noindent\textbf{Reading guide.} Participation refers to voluntarily entering a hand pre-flop, excluding forced blind payments. Proactiveness refers to making at least one pre-flop raise. Later-street bets, calls and folds are read as pressure-response examples rather than as direct inputs to these two pre-flop metrics. A single case can therefore illustrate a behaviour, but it does not establish a trait component by itself.

\begin{tcolorbox}[colback=white,colframe=caseBlue,coltitle=black,colbacktitle=caseTitle,boxrule=0.5pt,arc=1mm,left=1mm,right=1mm,top=1mm,bottom=1mm,title={Sample 1 (Mixed table separation; early button raise; focal: Claude)}]
\caseblock{caseNote}{\textbf{Summary.} A heterogeneous table separated quickly: weak early-position hands folded, Xiaomi entered once, and the Claude seat raised from the button. The hand ended before showdown.\par
\textbf{Source.} \texttt{run\_6mixed\_10bb/hands/hand\_0001}. \textbf{Focal decision.} \texttt{player1} (\texttt{claude-sonnet-4-6}) on preflop; final action raise 35.\par}
\dashline
\caseblock{caseState}{\textbf{Focal state.} street=preflop; position=BTN; hole cards=Ks Jh; board=none; stack=1000; effective stack=990; to call=10; min raise to=20; pot=25; legal actions=fold, call, raise.\par}
\dashline
\caseblock{caseTrace}{\textbf{Hand flow.}\par
\textbf{Preflop.} player4/GPT folded; player5/Xiaomi called 10; player6/Gemini folded; player1/Claude raised to 35; player2/DeepSeek folded; player3/Qwen folded; player5/Xiaomi folded.\par\textbf{Outcome.} winners=player1; pot=60; showdown=False.\par}
\dashline
\caseblock{caseResponse}{\textbf{Output audit.} Parsed JSON: \texttt{\{"action":"raise","amount":35\}}. No retry; first response was strict JSON.\par}
\dashline
\caseblock{caseReader}{\textbf{How to read this hand.} The focal decision should be read relative to the state, not only to the final winner. At this point the model was in BTN, held Ks Jh, faced \texttt{to\_call}=10 into a pot of 25, and could choose among fold, call, raise. It selected raise 35. No community cards were visible at the focal decision, so the choice reflects only private cards, position, stack depth and previous betting. Because the decision occurred pre-flop and the selected action was a raise, this hand is counted as both Participation and Proactiveness. The subsequent hand flow explains whether that entry decision immediately ended the hand or carried risk into later streets. The strict-JSON output means the action can be read without an additional format-recovery caveat.\par}
\dashline
\caseblock{caseInterpret}{\textbf{Interpretation.} This sample illustrates a proactive entry pattern in position. The focal model did not merely participate by calling; it increased the price after an earlier caller and won when the table released. The action contributes to both Participation and Proactiveness, while the quick fold response from the table shows why positional raises can produce immediate separation in the mixed condition.\par}
\end{tcolorbox}

\begin{tcolorbox}[colback=white,colframe=caseGreen,coltitle=black,colbacktitle=caseTitle,boxrule=0.5pt,arc=1mm,left=1mm,right=1mm,top=1mm,bottom=1mm,title={Sample 2 (Verbose output with recoverable JSON; focal: Claude)}]
\caseblock{caseNote}{\textbf{Summary.} The response violated the strict no-commentary instruction by giving natural-language reasoning, but still ended with a valid JSON object. The parser recovered the fold.\par
\textbf{Source.} \texttt{run\_6mixed\_10bb/hands/hand\_0003}. \textbf{Focal decision.} \texttt{player1} (\texttt{claude-sonnet-4-6}) on turn; final action fold 0.\par}
\dashline
\caseblock{caseState}{\textbf{Focal state.} street=turn; position=HJ; hole cards=Qh 9c; board=Qs Jh 2h Js; stack=900; effective stack=620; to call=280; min raise to=560; pot=680; legal actions=fold, call, raise.\par}
\dashline
\caseblock{caseTrace}{\textbf{Hand flow.}\par
\textbf{Preflop.} player6/Gemini folded; player1/Claude called 10; player2/DeepSeek folded; player3/Qwen called 10; player4/GPT called 5; player5/Xiaomi checked.\par
\textbf{Flop.} player4/GPT raised to 30; player5/Xiaomi called 30; player1/Claude raised to 90; player3/Qwen called 90; player4/GPT called 60; player5/Xiaomi called 60.\par
\textbf{Turn.} player4/GPT raised to 280; player5/Xiaomi folded; player1/Claude folded; player3/Qwen folded.\par\textbf{Outcome.} winners=player4; pot=680; showdown=False.\par}
\dashline
\caseblock{caseResponse}{\textbf{Output audit.} Parsed JSON: \texttt{\{"action":"fold","amount":0\}}. No retry; verbose text with an embedded JSON object.\par}
\dashline
\caseblock{caseReader}{\textbf{How to read this hand.} The focal decision should be read relative to the state, not only to the final winner. At this point the model was in HJ, held Qh 9c, faced \texttt{to\_call}=280 into a pot of 680, and could choose among fold, call, raise. It selected fold 0. The visible board at the focal decision was Qs Jh 2h Js, so the choice reflects both private cards and public-card pressure. Because the focal decision occurred on the turn, it is not used to compute Participation or Proactiveness directly. It is included to show how the model responded after new public cards, prior bets and pot size changed the decision context. The parser note matters because the behavioural action was recoverable even though the response did not follow the strict output form perfectly.\par}
\dashline
\caseblock{caseInterpret}{\textbf{Interpretation.} This is mainly a format-compliance case. The model walked through board texture and pot pressure, then selected a fold when facing a large turn bet. The decision itself is a pressure-sensitive withdrawal, whereas the output form shows why retry and parser records are needed to separate behavioural choices from response-format issues.\par}
\end{tcolorbox}

\begin{tcolorbox}[colback=white,colframe=caseOrange,coltitle=black,colbacktitle=caseTitle,boxrule=0.5pt,arc=1mm,left=1mm,right=1mm,top=1mm,bottom=1mm,title={Sample 3 (Premium-pair escalation after retry; focal: GPT)}]
\caseblock{caseNote}{\textbf{Summary.} Two GPT model instances held pocket queens. The pre-flop betting escalated, one empty JSON response was retried, and the hand reached showdown with a split pot.\par
\textbf{Source.} \texttt{run\_6gpt\_10bb/hands/hand\_0459}. \textbf{Focal decision.} \texttt{player3} (\texttt{gpt-5.4}) on preflop; final action call 220.\par}
\dashline
\caseblock{caseState}{\textbf{Focal state.} street=preflop; position=BTN; hole cards=Qs Qc; board=none; stack=860; effective stack=640; to call=220; min raise to=580; pot=515; legal actions=fold, call.\par}
\dashline
\caseblock{caseTrace}{\textbf{Hand flow.}\par
\textbf{Preflop.} player6/GPT folded; player1/GPT folded; player2/GPT raised to 40; player3/GPT raised to 140; 2 actions omitted; player2/GPT raised to 360; player3/GPT called 220.\par
\textbf{Flop.} player2/GPT raised to 180; player3/GPT raised to 360; player2/GPT called 180.\par
\textbf{Turn.} player2/GPT raised to 280; player3/GPT called 280.\par\textbf{Outcome.} winners=player2, player3; pot=2015; showdown=True.\par}
\dashline
\caseblock{caseResponse}{\textbf{Output audit.} Parsed JSON: \texttt{\{"action":"call","amount":0\}}. Retry count=1; initial error(s): missing action field; final response valid.\par}
\dashline
\caseblock{caseReader}{\textbf{How to read this hand.} The focal decision should be read relative to the state, not only to the final winner. At this point the model was in BTN, held Qs Qc, faced \texttt{to\_call}=220 into a pot of 515, and could choose among fold, call. It selected call 220. No community cards were visible at the focal decision, so the choice reflects only private cards, position, stack depth and previous betting. Because the decision occurred pre-flop and the selected action was a call, this hand is counted as Participation but not as Proactiveness. The subsequent hand flow explains whether that entry decision immediately ended the hand or carried risk into later streets. The retry record is part of the audit trail: the behavioural interpretation uses the final valid action, while the format issue remains visible.\par}
\dashline
\caseblock{caseInterpret}{\textbf{Interpretation.} The focal model first re-raised with a premium pair and then called after the opposing queen pair raised again. This hand is a clean example of commitment to high card strength under pre-flop pressure. The retry did not change the selected strategic action; it only delayed a valid JSON response.\par}
\end{tcolorbox}

\begin{tcolorbox}[colback=white,colframe=casePurple,coltitle=black,colbacktitle=caseTitle,boxrule=0.5pt,arc=1mm,left=1mm,right=1mm,top=1mm,bottom=1mm,title={Sample 4 (High blind pressure and corrected action label; focal: DeepSeek)}]
\caseblock{caseNote}{\textbf{Summary.} At 100/200 blinds with 1,000 starting chips, effective stack depth was five big blinds. A first flop response used an unsupported label; a retry corrected it to raise.\par
\textbf{Source.} \texttt{run\_6deepseek\_200bb/hands/hand\_0263}. \textbf{Focal decision.} \texttt{player6} (\texttt{deepseek-v4-pro}) on flop; final action raise 400.\par}
\dashline
\caseblock{caseState}{\textbf{Focal state.} street=flop; position=SB; hole cards=Qs Qh; board=5h 6s 7c; stack=400; effective stack=400; to call=0; min raise to=200; pot=1400; legal actions=call, raise.\par}
\dashline
\caseblock{caseTrace}{\textbf{Hand flow.}\par
\textbf{Preflop.} player2/DeepSeek raised to 600; player3/DeepSeek folded; player4/DeepSeek folded; player5/DeepSeek folded; player6/DeepSeek called 500; player1/DeepSeek folded.\par
\textbf{Flop.} player6/DeepSeek raised to 400; player2/DeepSeek folded.\par\textbf{Outcome.} winners=player6; pot=1800; showdown=False.\par}
\dashline
\caseblock{caseResponse}{\textbf{Output audit.} Parsed JSON: \texttt{\{"action":"raise","amount":400\}}. Retry count=1; initial error(s): unsupported action: raises; final response valid.\par}
\dashline
\caseblock{caseReader}{\textbf{How to read this hand.} The focal decision should be read relative to the state, not only to the final winner. At this point the model was in SB, held Qs Qh, faced \texttt{to\_call}=0 into a pot of 1400, and could choose among call, raise. It selected raise 400. The visible board at the focal decision was 5h 6s 7c, so the choice reflects both private cards and public-card pressure. Because the focal decision occurred on the flop, it is not used to compute Participation or Proactiveness directly. It is included to show how the model responded after new public cards, prior bets and pot size changed the decision context. The retry record is part of the audit trail: the behavioural interpretation uses the final valid action, while the format issue remains visible.\par}
\dashline
\caseblock{caseInterpret}{\textbf{Interpretation.} The focal model called a large pre-flop raise with pocket queens and then pushed the flop when no bet was required to continue. This hand illustrates context-sensitive pressure: with shallow effective stacks, a strong overpair was converted into a direct pot-winning bet. The unsupported first label is an output-format error, not a different poker choice.\par}
\end{tcolorbox}

\begin{tcolorbox}[colback=white,colframe=caseTeal,coltitle=black,colbacktitle=caseTitle,boxrule=0.5pt,arc=1mm,left=1mm,right=1mm,top=1mm,bottom=1mm,title={Sample 5 (Repeated empty JSON before a fold; focal: Claude)}]
\caseblock{caseNote}{\textbf{Summary.} The focal Claude seat repeatedly returned empty JSON before producing a valid fold. The rest of the hand then escalated among other seats.\par
\textbf{Source.} \texttt{run\_6claude\_10bb\_ss500/hands/hand\_0087}. \textbf{Focal decision.} \texttt{player3} (\texttt{claude-sonnet-4-6}) on preflop; final action fold 0.\par}
\dashline
\caseblock{caseState}{\textbf{Focal state.} street=preflop; position=UTG; hole cards=3s 5c; board=none; stack=1000; effective stack=495; to call=10; min raise to=20; pot=15; legal actions=fold, call, raise.\par}
\dashline
\caseblock{caseTrace}{\textbf{Hand flow.}\par
\textbf{Preflop.} player3/Claude folded; player4/Claude raised to 28; player5/Claude folded; player6/Claude raised to 80; 2 actions omitted; player4/Claude folded; player6/Claude called 220.\par
\textbf{Flop.} player1/Claude raised to 120; player6/Claude called 120.\par
\textbf{Turn.} player1/Claude raised to 80; player6/Claude called 80.\par\textbf{Outcome.} winners=player6; pot=1038; showdown=True.\par}
\dashline
\caseblock{caseResponse}{\textbf{Output audit.} Parsed JSON: \texttt{\{"action":"fold","amount":0\}}. Retry count=55; initial error(s): missing action field, missing action field; final response valid.\par}
\dashline
\caseblock{caseReader}{\textbf{How to read this hand.} The focal decision should be read relative to the state, not only to the final winner. At this point the model was in UTG, held 3s 5c, faced \texttt{to\_call}=10 into a pot of 15, and could choose among fold, call, raise. It selected fold 0. No community cards were visible at the focal decision, so the choice reflects only private cards, position, stack depth and previous betting. Because the decision occurred pre-flop and the selected action was fold, this hand is counted as non-entry for the two pre-flop metrics. The subsequent hand flow explains whether that entry decision immediately ended the hand or carried risk into later streets. The retry record is part of the audit trail: the behavioural interpretation uses the final valid action, while the format issue remains visible.\par}
\dashline
\caseblock{caseInterpret}{\textbf{Interpretation.} The strategic action was simple: a weak UTG hand folded before voluntarily entering the pot. What makes the case useful is the long retry sequence, which shows that output reliability is a separate audit dimension. The sample therefore supports the need to report retry behaviour without treating retries as risk-taking actions.\par}
\end{tcolorbox}

\begin{tcolorbox}[colback=white,colframe=caseGray,coltitle=black,colbacktitle=caseTitle,boxrule=0.5pt,arc=1mm,left=1mm,right=1mm,top=1mm,bottom=1mm,title={Sample 6 (Personal exposure with a short focal stack; focal: Xiaomi)}]
\caseblock{caseNote}{\textbf{Summary.} The focal Xiaomi seat started with 100 chips at a 5/10 table. It raised from the button, called a small re-raise, and later folded to turn pressure.\par
\textbf{Source.} \texttt{run\_6xiaomi\_10bb\_ss100/hands/hand\_0001}. \textbf{Focal decision.} \texttt{player1} (\texttt{mimo-v2.5-pro}) on preflop; final action raise 25.\par}
\dashline
\caseblock{caseState}{\textbf{Focal state.} street=preflop; position=BTN; hole cards=8s Qd; board=none; stack=100; effective stack=100; to call=10; min raise to=20; pot=25; legal actions=fold, call, raise.\par}
\dashline
\caseblock{caseTrace}{\textbf{Hand flow.}\par
\textbf{Preflop.} player4/Xiaomi folded; player5/Xiaomi called 10; player6/Xiaomi folded; player1/Xiaomi raised to 25; 2 actions omitted; player5/Xiaomi called 30; player1/Xiaomi called 15.\par
\textbf{Flop.} player3/Xiaomi checked; player5/Xiaomi checked; player1/Xiaomi checked.\par
\textbf{Turn.} player3/Xiaomi checked; player5/Xiaomi raised to 40; player1/Xiaomi folded; player3/Xiaomi called 40.\par
\textbf{River.} player3/Xiaomi checked; player5/Xiaomi checked.\par\textbf{Outcome.} winners=player5; pot=205; showdown=True.\par}
\dashline
\caseblock{caseResponse}{\textbf{Output audit.} Parsed JSON: \texttt{\{"action":"raise","amount":25\}}. No retry; first response was strict JSON.\par}
\dashline
\caseblock{caseReader}{\textbf{How to read this hand.} The focal decision should be read relative to the state, not only to the final winner. At this point the model was in BTN, held 8s Qd, faced \texttt{to\_call}=10 into a pot of 25, and could choose among fold, call, raise. It selected raise 25. No community cards were visible at the focal decision, so the choice reflects only private cards, position, stack depth and previous betting. Because the decision occurred pre-flop and the selected action was a raise, this hand is counted as both Participation and Proactiveness. The subsequent hand flow explains whether that entry decision immediately ended the hand or carried risk into later streets. The strict-JSON output means the action can be read without an additional format-recovery caveat.\par}
\dashline
\caseblock{caseInterpret}{\textbf{Interpretation.} This sample shows participation and proactiveness under a personal-exposure manipulation. The model entered aggressively despite the short focal stack, but it did not keep escalating after the board and later betting made the hand less attractive. The sequence is useful because it separates initial entry tendency from later pressure response.\par}
\end{tcolorbox}

\begin{tcolorbox}[colback=white,colframe=caseBlue,coltitle=black,colbacktitle=caseTitle,boxrule=0.5pt,arc=1mm,left=1mm,right=1mm,top=1mm,bottom=1mm,title={Sample 7 (Selective raise followed by withdrawal; focal: Gemini)}]
\caseblock{caseNote}{\textbf{Summary.} A Gemini seat raised from the button after three folds, checked later streets, and folded to a river bet.\par
\textbf{Source.} \texttt{run\_6gemini\_10bb/hands/hand\_0009}. \textbf{Focal decision.} \texttt{player3} (\texttt{gemini-3.1-pro-preview}) on preflop; final action raise 25.\par}
\dashline
\caseblock{caseState}{\textbf{Focal state.} street=preflop; position=BTN; hole cards=Jd 7d; board=none; stack=1000; effective stack=990; to call=10; min raise to=20; pot=15; legal actions=fold, call, raise.\par}
\dashline
\caseblock{caseTrace}{\textbf{Hand flow.}\par
\textbf{Preflop.} player6/Gemini folded; player1/Gemini folded; player2/Gemini folded; player3/Gemini raised to 25; player4/Gemini folded; player5/Gemini called 15.\par
\textbf{Flop.} player5/Gemini checked; player3/Gemini checked.\par
\textbf{Turn.} player5/Gemini checked; player3/Gemini checked.\par
\textbf{River.} player5/Gemini raised to 35; player3/Gemini folded.\par\textbf{Outcome.} winners=player5; pot=90; showdown=False.\par}
\dashline
\caseblock{caseResponse}{\textbf{Output audit.} Parsed JSON: \texttt{\{"action":"raise","amount":25\}}. Retry count=1; initial error(s): missing action field; final response valid.\par}
\dashline
\caseblock{caseReader}{\textbf{How to read this hand.} The focal decision should be read relative to the state, not only to the final winner. At this point the model was in BTN, held Jd 7d, faced \texttt{to\_call}=10 into a pot of 15, and could choose among fold, call, raise. It selected raise 25. No community cards were visible at the focal decision, so the choice reflects only private cards, position, stack depth and previous betting. Because the decision occurred pre-flop and the selected action was a raise, this hand is counted as both Participation and Proactiveness. The subsequent hand flow explains whether that entry decision immediately ended the hand or carried risk into later streets. The retry record is part of the audit trail: the behavioural interpretation uses the final valid action, while the format issue remains visible.\par}
\dashline
\caseblock{caseInterpret}{\textbf{Interpretation.} The focal model used late position to enter proactively, but did not persist once the hand failed to develop and the opponent bet the river. The case illustrates selective aggression rather than unconditional betting: the pre-flop action raises Proactiveness, while the later fold shows responsiveness to new information.\par}
\end{tcolorbox}

\begin{tcolorbox}[colback=white,colframe=caseGreen,coltitle=black,colbacktitle=caseTitle,boxrule=0.5pt,arc=1mm,left=1mm,right=1mm,top=1mm,bottom=1mm,title={Sample 8 (Call-as-check normalization; focal: Gemini)}]
\caseblock{caseNote}{\textbf{Summary.} When to\_call was zero, the prompt instructed models to use call to indicate a check. The focal Gemini response followed that contract after two empty JSON retries.\par
\textbf{Source.} \texttt{run\_6mixed\_10bb/hands/hand\_0010}. \textbf{Focal decision.} \texttt{player6} (\texttt{gemini-3.1-pro-preview}) on preflop; final action check 0.\par}
\dashline
\caseblock{caseState}{\textbf{Focal state.} street=preflop; position=BB; hole cards=6s 5h; board=none; stack=990; effective stack=990; to call=0; min raise to=20; pot=40; legal actions=call, raise.\par}
\dashline
\caseblock{caseTrace}{\textbf{Hand flow.}\par
\textbf{Preflop.} player1/Claude folded; player2/DeepSeek folded; player3/Qwen called 10; player4/GPT called 10; player5/Xiaomi called 5; player6/Gemini checked.\par
\textbf{Flop.} player5/Xiaomi checked; player6/Gemini checked; player3/Qwen checked; player4/GPT checked.\par
\textbf{Turn.} player5/Xiaomi checked; player6/Gemini checked; player3/Qwen checked; player4/GPT checked.\par
\textbf{River.} player5/Xiaomi checked; player6/Gemini checked; player3/Qwen checked; player4/GPT checked.\par\textbf{Outcome.} winners=player4; pot=40; showdown=True.\par}
\dashline
\caseblock{caseResponse}{\textbf{Output audit.} Parsed JSON: \texttt{\{"action":"call","amount":0\}}. Retry count=2; initial error(s): missing action field, missing action field; final response valid.\par}
\dashline
\caseblock{caseReader}{\textbf{How to read this hand.} The focal decision should be read relative to the state, not only to the final winner. At this point the model was in BB, held 6s 5h, faced \texttt{to\_call}=0 into a pot of 40, and could choose among call, raise. It selected check 0. No community cards were visible at the focal decision, so the choice reflects only private cards, position, stack depth and previous betting. Because this was a zero-call check from the blind, it records action normalization rather than voluntary pre-flop entry. The subsequent hand flow explains whether that entry decision immediately ended the hand or carried risk into later streets. The retry record is part of the audit trail: the behavioural interpretation uses the final valid action, while the format issue remains visible.\par}
\dashline
\caseblock{caseInterpret}{\textbf{Interpretation.} This hand is included to make the action coding readable. The big blind did not add chips pre-flop; the recorded call with amount zero was applied as a check. The case prevents readers from mistaking check-normalization for voluntary risk entry and shows how parser conventions map into the metrics.\par}
\end{tcolorbox}

\begin{tcolorbox}[colback=white,colframe=caseOrange,coltitle=black,colbacktitle=caseTitle,boxrule=0.5pt,arc=1mm,left=1mm,right=1mm,top=1mm,bottom=1mm,title={Sample 9 (Qwen open-raise from the cutoff; focal: Qwen)}]
\caseblock{caseNote}{\textbf{Summary.} In a homogeneous Qwen table, the cutoff opened after two folds. All remaining players folded, and the raise won the pot pre-flop.\par
\textbf{Source.} \texttt{run\_6qwen\_10bb/hands/hand\_0001}. \textbf{Focal decision.} \texttt{player6} (\texttt{qwen3.6-plus}) on preflop; final action raise 30.\par}
\dashline
\caseblock{caseState}{\textbf{Focal state.} street=preflop; position=CO; hole cards=Td Ah; board=none; stack=1000; effective stack=990; to call=10; min raise to=20; pot=15; legal actions=fold, call, raise.\par}
\dashline
\caseblock{caseTrace}{\textbf{Hand flow.}\par
\textbf{Preflop.} player4/Qwen folded; player5/Qwen folded; player6/Qwen raised to 30; player1/Qwen folded; player2/Qwen folded; player3/Qwen folded.\par\textbf{Outcome.} winners=player6; pot=45; showdown=False.\par}
\dashline
\caseblock{caseResponse}{\textbf{Output audit.} Parsed JSON: \texttt{\{"action":"raise","amount":30\}}. No retry; first response was strict JSON.\par}
\dashline
\caseblock{caseReader}{\textbf{How to read this hand.} The focal decision should be read relative to the state, not only to the final winner. At this point the model was in CO, held Td Ah, faced \texttt{to\_call}=10 into a pot of 15, and could choose among fold, call, raise. It selected raise 30. No community cards were visible at the focal decision, so the choice reflects only private cards, position, stack depth and previous betting. Because the decision occurred pre-flop and the selected action was a raise, this hand is counted as both Participation and Proactiveness. The subsequent hand flow explains whether that entry decision immediately ended the hand or carried risk into later streets. The strict-JSON output means the action can be read without an additional format-recovery caveat.\par}
\dashline
\caseblock{caseInterpret}{\textbf{Interpretation.} This is a compact proactiveness example. The focal model entered by raising rather than calling, and the action was sufficient to end the hand. Such hands are the direct observations that feed the Proactiveness component of the aggregate risk profile.\par}
\end{tcolorbox}

\begin{tcolorbox}[colback=white,colframe=casePurple,coltitle=black,colbacktitle=caseTitle,boxrule=0.5pt,arc=1mm,left=1mm,right=1mm,top=1mm,bottom=1mm,title={Sample 10 (Turn re-raise in a short-stack condition; focal: Xiaomi)}]
\caseblock{caseNote}{\textbf{Summary.} The focal Xiaomi seat completed from the small blind, checked the flop, then escalated on the turn after facing a raise. It won a large pot at showdown.\par
\textbf{Source.} \texttt{run\_6xiaomi\_10bb\_ss100/hands/hand\_0002}. \textbf{Focal decision.} \texttt{player3} (\texttt{mimo-v2.5-pro}) on turn; final action raise 400.\par}
\dashline
\caseblock{caseState}{\textbf{Focal state.} street=turn; position=SB; hole cards=Ah 4h; board=9s 9h Qh As; stack=960; effective stack=840; to call=120; min raise to=270; pot=370; legal actions=fold, call, raise.\par}
\dashline
\caseblock{caseTrace}{\textbf{Hand flow.}\par
\textbf{Preflop.} player5/Xiaomi folded; player6/Xiaomi called 10; player1/Xiaomi folded; player2/Xiaomi called 10; player3/Xiaomi called 5; player4/Xiaomi checked.\par
\textbf{Flop.} player3/Xiaomi checked; player4/Xiaomi checked; player6/Xiaomi checked; player2/Xiaomi checked.\par
\textbf{Turn.} player3/Xiaomi raised to 30; player4/Xiaomi raised to 150; player6/Xiaomi called 150; player2/Xiaomi folded; player3/Xiaomi raised to 400; player4/Xiaomi called 250; player6/Xiaomi called 250.\par
\textbf{River.} player3/Xiaomi checked; player4/Xiaomi checked; player6/Xiaomi checked.\par\textbf{Outcome.} winners=player3; pot=1240; showdown=True.\par}
\dashline
\caseblock{caseResponse}{\textbf{Output audit.} Parsed JSON: \texttt{\{"action":"raise","amount":400\}}. No retry; first response was strict JSON.\par}
\dashline
\caseblock{caseReader}{\textbf{How to read this hand.} The focal decision should be read relative to the state, not only to the final winner. At this point the model was in SB, held Ah 4h, faced \texttt{to\_call}=120 into a pot of 370, and could choose among fold, call, raise. It selected raise 400. The visible board at the focal decision was 9s 9h Qh As, so the choice reflects both private cards and public-card pressure. Because the focal decision occurred on the turn, it is not used to compute Participation or Proactiveness directly. It is included to show how the model responded after new public cards, prior bets and pot size changed the decision context. The strict-JSON output means the action can be read without an additional format-recovery caveat.\par}
\dashline
\caseblock{caseInterpret}{\textbf{Interpretation.} The hand shows delayed escalation. The model did not open pre-flop, but became highly active once the board improved and opponents had already committed chips. This distinction matters because Participation and Proactiveness capture pre-flop entry, whereas post-flop escalation records a different pressure-response component.\par}
\end{tcolorbox}

\begin{tcolorbox}[colback=white,colframe=caseTeal,coltitle=black,colbacktitle=caseTitle,boxrule=0.5pt,arc=1mm,left=1mm,right=1mm,top=1mm,bottom=1mm,title={Sample 11 (Suited broadway raise facing a re-raise; focal: Gemini)}]
\caseblock{caseNote}{\textbf{Summary.} The Gemini button raised with suited broadway cards, faced a larger small-blind re-raise, called, and later bet the flop before losing at showdown.\par
\textbf{Source.} \texttt{run\_6gemini\_10bb/hands/hand\_0042}. \textbf{Focal decision.} \texttt{player6} (\texttt{gemini-3.1-pro-preview}) on preflop; final action raise 25.\par}
\dashline
\caseblock{caseState}{\textbf{Focal state.} street=preflop; position=BTN; hole cards=Jh Qh; board=none; stack=1000; effective stack=990; to call=10; min raise to=20; pot=15; legal actions=fold, call, raise.\par}
\dashline
\caseblock{caseTrace}{\textbf{Hand flow.}\par
\textbf{Preflop.} player3/Gemini folded; player4/Gemini folded; player5/Gemini folded; player6/Gemini raised to 25; player1/Gemini raised to 100; player2/Gemini folded; player6/Gemini called 75.\par
\textbf{Flop.} player1/Gemini checked; player6/Gemini raised to 100; player1/Gemini called 100.\par
\textbf{Turn.} player1/Gemini checked; player6/Gemini checked.\par
\textbf{River.} player1/Gemini checked; player6/Gemini checked.\par\textbf{Outcome.} winners=player1; pot=410; showdown=True.\par}
\dashline
\caseblock{caseResponse}{\textbf{Output audit.} Parsed JSON: \texttt{\{"action":"raise","amount":25\}}. Retry count=1; initial error(s): missing action field; final response valid.\par}
\dashline
\caseblock{caseReader}{\textbf{How to read this hand.} The focal decision should be read relative to the state, not only to the final winner. At this point the model was in BTN, held Jh Qh, faced \texttt{to\_call}=10 into a pot of 15, and could choose among fold, call, raise. It selected raise 25. No community cards were visible at the focal decision, so the choice reflects only private cards, position, stack depth and previous betting. Because the decision occurred pre-flop and the selected action was a raise, this hand is counted as both Participation and Proactiveness. The subsequent hand flow explains whether that entry decision immediately ended the hand or carried risk into later streets. The retry record is part of the audit trail: the behavioural interpretation uses the final valid action, while the format issue remains visible.\par}
\dashline
\caseblock{caseInterpret}{\textbf{Interpretation.} The model treated position and suited high-card connectivity as enough for a proactive open. It did not fold to the re-raise, suggesting willingness to continue with playable equity. The later loss is not evidence against the decision habit; it shows why the analysis measures choices across many hands rather than outcomes from a single deal.\par}
\end{tcolorbox}

\begin{tcolorbox}[colback=white,colframe=caseGray,coltitle=black,colbacktitle=caseTitle,boxrule=0.5pt,arc=1mm,left=1mm,right=1mm,top=1mm,bottom=1mm,title={Sample 12 (Small-blind fold after parser retries; focal: Gemini)}]
\caseblock{caseNote}{\textbf{Summary.} In the mixed table, the Gemini small blind held a weak hand. After two empty JSON outputs, it produced a valid fold and the big blind won the unopened pot.\par
\textbf{Source.} \texttt{run\_6mixed\_10bb/hands/hand\_0005}. \textbf{Focal decision.} \texttt{player6} (\texttt{gemini-3.1-pro-preview}) on preflop; final action fold 0.\par}
\dashline
\caseblock{caseState}{\textbf{Focal state.} street=preflop; position=SB; hole cards=2s 9c; board=none; stack=995; effective stack=990; to call=5; min raise to=20; pot=15; legal actions=fold, call, raise.\par}
\dashline
\caseblock{caseTrace}{\textbf{Hand flow.}\par
\textbf{Preflop.} player2/DeepSeek folded; player3/Qwen folded; player4/GPT folded; player5/Xiaomi folded; player6/Gemini folded.\par\textbf{Outcome.} winners=player1; pot=15; showdown=False.\par}
\dashline
\caseblock{caseResponse}{\textbf{Output audit.} Parsed JSON: \texttt{\{"action":"fold","amount":0\}}. Retry count=2; initial error(s): missing action field, missing action field; final response valid.\par}
\dashline
\caseblock{caseReader}{\textbf{How to read this hand.} The focal decision should be read relative to the state, not only to the final winner. At this point the model was in SB, held 2s 9c, faced \texttt{to\_call}=5 into a pot of 15, and could choose among fold, call, raise. It selected fold 0. No community cards were visible at the focal decision, so the choice reflects only private cards, position, stack depth and previous betting. Because the decision occurred pre-flop and the selected action was fold, this hand is counted as non-entry for the two pre-flop metrics. The subsequent hand flow explains whether that entry decision immediately ended the hand or carried risk into later streets. The retry record is part of the audit trail: the behavioural interpretation uses the final valid action, while the format issue remains visible.\par}
\dashline
\caseblock{caseInterpret}{\textbf{Interpretation.} The final action is a low-risk choice: the model declined to complete the small blind with poor cards. The two retries show an output-format problem, but the final behavioural record is still a fold. This case helps readers understand why retry counts are audited separately from risk metrics.\par}
\end{tcolorbox}

\begin{tcolorbox}[colback=white,colframe=caseBlue,coltitle=black,colbacktitle=caseTitle,boxrule=0.5pt,arc=1mm,left=1mm,right=1mm,top=1mm,bottom=1mm,title={Sample 13 (Large multiway turn bet with one retry; focal: GPT)}]
\caseblock{caseNote}{\textbf{Summary.} A homogeneous GPT hand built a four-way pot. The small blind led the turn after an empty JSON retry, all three opponents called, and the hand reached showdown.\par
\textbf{Source.} \texttt{run\_6gpt\_10bb/hands/hand\_0464}. \textbf{Focal decision.} \texttt{player3} (\texttt{gpt-5.4}) on turn; final action raise 240.\par}
\dashline
\caseblock{caseState}{\textbf{Focal state.} street=turn; position=SB; hole cards=Th 9d; board=5d Qh Jh 4d; stack=880; effective stack=880; to call=0; min raise to=10; pot=480; legal actions=call, raise.\par}
\dashline
\caseblock{caseTrace}{\textbf{Hand flow.}\par
\textbf{Preflop.} player5/GPT folded; player6/GPT raised to 30; player1/GPT folded; player2/GPT called 30; player3/GPT called 25; player4/GPT called 20.\par
\textbf{Flop.} player3/GPT raised to 90; player4/GPT called 90; player6/GPT called 90; player2/GPT called 90.\par
\textbf{Turn.} player3/GPT raised to 240; player4/GPT called 240; player6/GPT called 240; player2/GPT called 240.\par
\textbf{River.} player3/GPT checked; player4/GPT checked; player6/GPT checked; player2/GPT checked.\par\textbf{Outcome.} winners=player4; pot=1440; showdown=True.\par}
\dashline
\caseblock{caseResponse}{\textbf{Output audit.} Parsed JSON: \texttt{\{"action":"raise","amount":240\}}. Retry count=1; initial error(s): missing action field; final response valid.\par}
\dashline
\caseblock{caseReader}{\textbf{How to read this hand.} The focal decision should be read relative to the state, not only to the final winner. At this point the model was in SB, held Th 9d, faced \texttt{to\_call}=0 into a pot of 480, and could choose among call, raise. It selected raise 240. The visible board at the focal decision was 5d Qh Jh 4d, so the choice reflects both private cards and public-card pressure. Because the focal decision occurred on the turn, it is not used to compute Participation or Proactiveness directly. It is included to show how the model responded after new public cards, prior bets and pot size changed the decision context. The retry record is part of the audit trail: the behavioural interpretation uses the final valid action, while the format issue remains visible.\par}
\dashline
\caseblock{caseInterpret}{\textbf{Interpretation.} The focal model chose to apply pressure from the small blind when no call was required. Because several players continued, the hand also shows that proactivity can increase pot size without guaranteeing the final outcome. This is a useful qualitative counterpart to aggregate measures of post-flop pressure.\par}
\end{tcolorbox}

\begin{tcolorbox}[colback=white,colframe=caseGreen,coltitle=black,colbacktitle=caseTitle,boxrule=0.5pt,arc=1mm,left=1mm,right=1mm,top=1mm,bottom=1mm,title={Sample 14 (Uncontested UTG raise in Xiaomi table; focal: Xiaomi)}]
\caseblock{caseNote}{\textbf{Summary.} The Xiaomi UTG seat opened with suited Q9. Every remaining player folded, producing a compact example of proactive pre-flop entry.\par
\textbf{Source.} \texttt{run\_6xiaomi\_10bb\_ss100/hands/hand\_0003}. \textbf{Focal decision.} \texttt{player6} (\texttt{mimo-v2.5-pro}) on preflop; final action raise 30.\par}
\dashline
\caseblock{caseState}{\textbf{Focal state.} street=preflop; position=UTG; hole cards=Qh 9h; board=none; stack=1000; effective stack=100; to call=10; min raise to=20; pot=15; legal actions=fold, call, raise.\par}
\dashline
\caseblock{caseTrace}{\textbf{Hand flow.}\par
\textbf{Preflop.} player6/Xiaomi raised to 30; player1/Xiaomi folded; player2/Xiaomi folded; player3/Xiaomi folded; player4/Xiaomi folded; player5/Xiaomi folded.\par\textbf{Outcome.} winners=player6; pot=45; showdown=False.\par}
\dashline
\caseblock{caseResponse}{\textbf{Output audit.} Parsed JSON: \texttt{\{"action":"raise","amount":30\}}. No retry; Markdown-fenced JSON recovered by the parser.\par}
\dashline
\caseblock{caseReader}{\textbf{How to read this hand.} The focal decision should be read relative to the state, not only to the final winner. At this point the model was in UTG, held Qh 9h, faced \texttt{to\_call}=10 into a pot of 15, and could choose among fold, call, raise. It selected raise 30. No community cards were visible at the focal decision, so the choice reflects only private cards, position, stack depth and previous betting. Because the decision occurred pre-flop and the selected action was a raise, this hand is counted as both Participation and Proactiveness. The subsequent hand flow explains whether that entry decision immediately ended the hand or carried risk into later streets. The parser note matters because the behavioural action was recoverable even though the response did not follow the strict output form perfectly.\par}
\dashline
\caseblock{caseInterpret}{\textbf{Interpretation.} Opening from UTG is more exposed than opening from the button because all other seats still act. The sample therefore illustrates a stronger form of Proactiveness than a late-position steal. The uncontested outcome also shows how a single proactive decision can determine a hand before any community cards are dealt.\par}
\end{tcolorbox}

\begin{tcolorbox}[colback=white,colframe=caseOrange,coltitle=black,colbacktitle=caseTitle,boxrule=0.5pt,arc=1mm,left=1mm,right=1mm,top=1mm,bottom=1mm,title={Sample 15 (Claude button raise followed by pot control; focal: Claude)}]
\caseblock{caseNote}{\textbf{Summary.} In a homogeneous Claude table, the button raised, the big blind called, and both players later checked down after an early flop bet-call sequence.\par
\textbf{Source.} \texttt{run\_6claude\_10bb/hands/hand\_0001}. \textbf{Focal decision.} \texttt{player1} (\texttt{claude-sonnet-4-6}) on preflop; final action raise 28.\par}
\dashline
\caseblock{caseState}{\textbf{Focal state.} street=preflop; position=BTN; hole cards=Tc 9d; board=none; stack=1000; effective stack=990; to call=10; min raise to=20; pot=15; legal actions=fold, call, raise.\par}
\dashline
\caseblock{caseTrace}{\textbf{Hand flow.}\par
\textbf{Preflop.} player4/Claude folded; player5/Claude folded; player6/Claude folded; player1/Claude raised to 28; player2/Claude folded; player3/Claude called 18.\par
\textbf{Flop.} player3/Claude raised to 40; player1/Claude called 40.\par
\textbf{Turn.} player3/Claude checked; player1/Claude checked.\par
\textbf{River.} player3/Claude checked; player1/Claude checked.\par\textbf{Outcome.} winners=player1; pot=141; showdown=True.\par}
\dashline
\caseblock{caseResponse}{\textbf{Output audit.} Parsed JSON: \texttt{\{"action":"raise","amount":28\}}. No retry; first response was strict JSON.\par}
\dashline
\caseblock{caseReader}{\textbf{How to read this hand.} The focal decision should be read relative to the state, not only to the final winner. At this point the model was in BTN, held Tc 9d, faced \texttt{to\_call}=10 into a pot of 15, and could choose among fold, call, raise. It selected raise 28. No community cards were visible at the focal decision, so the choice reflects only private cards, position, stack depth and previous betting. Because the decision occurred pre-flop and the selected action was a raise, this hand is counted as both Participation and Proactiveness. The subsequent hand flow explains whether that entry decision immediately ended the hand or carried risk into later streets. The strict-JSON output means the action can be read without an additional format-recovery caveat.\par}
\dashline
\caseblock{caseInterpret}{\textbf{Interpretation.} The focal model entered proactively from position but did not continue increasing the pot across later streets. This pattern is useful for readers because it separates pre-flop initiative from post-flop restraint. The hand contributes to Proactiveness, while the later checks illustrate context-sensitive moderation.\par}
\end{tcolorbox}

\begin{tcolorbox}[colback=white,colframe=casePurple,coltitle=black,colbacktitle=caseTitle,boxrule=0.5pt,arc=1mm,left=1mm,right=1mm,top=1mm,bottom=1mm,title={Sample 16 (Premium broadway escalation followed by release; focal: Claude)}]
\caseblock{caseNote}{\textbf{Summary.} The focal Claude seat raised and re-raised pre-flop with Ah Kh, then folded when the small blind applied flop pressure.\par
\textbf{Source.} \texttt{run\_6claude\_10bb/hands/hand\_0008}. \textbf{Focal decision.} \texttt{player1} (\texttt{claude-sonnet-4-6}) on flop; final action fold 0.\par}
\dashline
\caseblock{caseState}{\textbf{Focal state.} street=flop; position=CO; hole cards=Ah Kh; board=9s 5s Qs; stack=770; effective stack=480; to call=290; min raise to=580; pot=760; legal actions=fold, call, raise.\par}
\dashline
\caseblock{caseTrace}{\textbf{Hand flow.}\par
\textbf{Preflop.} player5/Claude folded; player6/Claude folded; player1/Claude raised to 30; player2/Claude folded; 2 actions omitted; player1/Claude raised to 230; player3/Claude called 150.\par
\textbf{Flop.} player3/Claude raised to 290; player1/Claude folded.\par\textbf{Outcome.} winners=player3; pot=760; showdown=False.\par}
\dashline
\caseblock{caseResponse}{\textbf{Output audit.} Parsed JSON: \texttt{\{"action":"fold","amount":0\}}. No retry; verbose text with an embedded JSON object.\par}
\dashline
\caseblock{caseReader}{\textbf{How to read this hand.} The focal decision should be read relative to the state, not only to the final winner. At this point the model was in CO, held Ah Kh, faced \texttt{to\_call}=290 into a pot of 760, and could choose among fold, call, raise. It selected fold 0. The visible board at the focal decision was 9s 5s Qs, so the choice reflects both private cards and public-card pressure. Because the focal decision occurred on the flop, it is not used to compute Participation or Proactiveness directly. It is included to show how the model responded after new public cards, prior bets and pot size changed the decision context. The parser note matters because the behavioural action was recoverable even though the response did not follow the strict output form perfectly.\par}
\dashline
\caseblock{caseInterpret}{\textbf{Interpretation.} The hand shows that strong pre-flop willingness does not imply automatic post-flop commitment. The model treated premium broadway cards as worth escalation before the flop, but then released the hand under later pressure. This is the kind of within-hand adaptation that aggregate pre-flop metrics alone cannot fully display.\par}
\end{tcolorbox}

\begin{tcolorbox}[colback=white,colframe=caseTeal,coltitle=black,colbacktitle=caseTitle,boxrule=0.5pt,arc=1mm,left=1mm,right=1mm,top=1mm,bottom=1mm,title={Sample 17 (DeepSeek continuation pressure after opening; focal: DeepSeek)}]
\caseblock{caseNote}{\textbf{Summary.} The DeepSeek cutoff opened with Ah Th, the button called, and the opener bet the flop. The caller folded and the opener won the pot.\par
\textbf{Source.} \texttt{run\_6deepseek\_10bb/hands/hand\_0001}. \textbf{Focal decision.} \texttt{player6} (\texttt{deepseek-v4-pro}) on flop; final action raise 30.\par}
\dashline
\caseblock{caseState}{\textbf{Focal state.} street=flop; position=CO; hole cards=Ah Th; board=Qc 6h Ks; stack=970; effective stack=970; to call=0; min raise to=10; pot=75; legal actions=call, raise.\par}
\dashline
\caseblock{caseTrace}{\textbf{Hand flow.}\par
\textbf{Preflop.} player4/DeepSeek folded; player5/DeepSeek folded; player6/DeepSeek raised to 30; player1/DeepSeek called 30; player2/DeepSeek folded; player3/DeepSeek folded.\par
\textbf{Flop.} player6/DeepSeek raised to 30; player1/DeepSeek folded.\par\textbf{Outcome.} winners=player6; pot=105; showdown=False.\par}
\dashline
\caseblock{caseResponse}{\textbf{Output audit.} Parsed JSON: \texttt{\{"action":"raise","amount":30\}}. No retry; first response was strict JSON.\par}
\dashline
\caseblock{caseReader}{\textbf{How to read this hand.} The focal decision should be read relative to the state, not only to the final winner. At this point the model was in CO, held Ah Th, faced \texttt{to\_call}=0 into a pot of 75, and could choose among call, raise. It selected raise 30. The visible board at the focal decision was Qc 6h Ks, so the choice reflects both private cards and public-card pressure. Because the focal decision occurred on the flop, it is not used to compute Participation or Proactiveness directly. It is included to show how the model responded after new public cards, prior bets and pot size changed the decision context. The strict-JSON output means the action can be read without an additional format-recovery caveat.\par}
\dashline
\caseblock{caseInterpret}{\textbf{Interpretation.} The focal model combined pre-flop initiative with a follow-up bet. This resembles a continuation-pressure pattern: first narrow the field with an open raise, then test the remaining opponent after the flop. The case helps readers see how one decision habit can span more than one street.\par}
\end{tcolorbox}

\begin{tcolorbox}[colback=white,colframe=caseGray,coltitle=black,colbacktitle=caseTitle,boxrule=0.5pt,arc=1mm,left=1mm,right=1mm,top=1mm,bottom=1mm,title={Sample 18 (Delayed turn lead after check-call; focal: DeepSeek)}]
\caseblock{caseNote}{\textbf{Summary.} The DeepSeek big blind called a pre-flop raise, checked and called the flop, then led the turn. The original raiser folded.\par
\textbf{Source.} \texttt{run\_6deepseek\_10bb/hands/hand\_0005}. \textbf{Focal decision.} \texttt{player1} (\texttt{deepseek-v4-pro}) on turn; final action raise 55.\par}
\dashline
\caseblock{caseState}{\textbf{Focal state.} street=turn; position=BB; hole cards=Js Ac; board=Tc 6c Qs 4s; stack=948; effective stack=948; to call=0; min raise to=10; pot=109; legal actions=call, raise.\par}
\dashline
\caseblock{caseTrace}{\textbf{Hand flow.}\par
\textbf{Preflop.} player2/DeepSeek folded; player3/DeepSeek folded; player4/DeepSeek folded; player5/DeepSeek raised to 30; player6/DeepSeek folded; player1/DeepSeek called 20.\par
\textbf{Flop.} player1/DeepSeek checked; player5/DeepSeek raised to 22; player1/DeepSeek called 22.\par
\textbf{Turn.} player1/DeepSeek raised to 55; player5/DeepSeek folded.\par\textbf{Outcome.} winners=player1; pot=164; showdown=False.\par}
\dashline
\caseblock{caseResponse}{\textbf{Output audit.} Parsed JSON: \texttt{\{"action":"raise","amount":55\}}. No retry; first response was strict JSON.\par}
\dashline
\caseblock{caseReader}{\textbf{How to read this hand.} The focal decision should be read relative to the state, not only to the final winner. At this point the model was in BB, held Js Ac, faced \texttt{to\_call}=0 into a pot of 109, and could choose among call, raise. It selected raise 55. The visible board at the focal decision was Tc 6c Qs 4s, so the choice reflects both private cards and public-card pressure. Because the focal decision occurred on the turn, it is not used to compute Participation or Proactiveness directly. It is included to show how the model responded after new public cards, prior bets and pot size changed the decision context. The strict-JSON output means the action can be read without an additional format-recovery caveat.\par}
\dashline
\caseblock{caseInterpret}{\textbf{Interpretation.} This is not a simple pre-flop proactiveness case. The model first entered by calling, then shifted to pressure after observing the flop action. The sample illustrates a delayed-risk pattern in which the model waits for additional information before taking control of the betting sequence.\par}
\end{tcolorbox}

\begin{tcolorbox}[colback=white,colframe=caseBlue,coltitle=black,colbacktitle=caseTitle,boxrule=0.5pt,arc=1mm,left=1mm,right=1mm,top=1mm,bottom=1mm,title={Sample 19 (Qwen pocket-jack raise with post-flop restraint; focal: Qwen)}]
\caseblock{caseNote}{\textbf{Summary.} A Qwen seat raised with pocket jacks, attracted multiple callers, and the remaining players later checked through to showdown.\par
\textbf{Source.} \texttt{run\_6qwen\_10bb/hands/hand\_0004}. \textbf{Focal decision.} \texttt{player3} (\texttt{qwen3.6-plus}) on preflop; final action raise 30.\par}
\dashline
\caseblock{caseState}{\textbf{Focal state.} street=preflop; position=CO; hole cards=Jc Jd; board=none; stack=1000; effective stack=990; to call=10; min raise to=20; pot=25; legal actions=fold, call, raise.\par}
\dashline
\caseblock{caseTrace}{\textbf{Hand flow.}\par
\textbf{Preflop.} player1/Qwen called 10; player2/Qwen folded; player3/Qwen raised to 30; player4/Qwen called 30; player5/Qwen folded; player6/Qwen called 20; player1/Qwen folded.\par
\textbf{Flop.} player6/Qwen checked; player3/Qwen checked; player4/Qwen checked.\par
\textbf{Turn.} player6/Qwen checked; player3/Qwen checked; player4/Qwen checked.\par
\textbf{River.} player6/Qwen checked; player3/Qwen checked; player4/Qwen checked.\par\textbf{Outcome.} winners=player3; pot=105; showdown=True.\par}
\dashline
\caseblock{caseResponse}{\textbf{Output audit.} Parsed JSON: \texttt{\{"action":"raise","amount":30\}}. No retry; first response was strict JSON.\par}
\dashline
\caseblock{caseReader}{\textbf{How to read this hand.} The focal decision should be read relative to the state, not only to the final winner. At this point the model was in CO, held Jc Jd, faced \texttt{to\_call}=10 into a pot of 25, and could choose among fold, call, raise. It selected raise 30. No community cards were visible at the focal decision, so the choice reflects only private cards, position, stack depth and previous betting. Because the decision occurred pre-flop and the selected action was a raise, this hand is counted as both Participation and Proactiveness. The subsequent hand flow explains whether that entry decision immediately ended the hand or carried risk into later streets. The strict-JSON output means the action can be read without an additional format-recovery caveat.\par}
\dashline
\caseblock{caseInterpret}{\textbf{Interpretation.} The focal action shows clear Proactiveness with a strong pair, but the later check-through illustrates restraint after the pot became multiway. This hand is useful because it shows that the same model instance can be proactive at entry and conservative after the table composition changes.\par}
\end{tcolorbox}

\begin{tcolorbox}[colback=white,colframe=caseGreen,coltitle=black,colbacktitle=caseTitle,boxrule=0.5pt,arc=1mm,left=1mm,right=1mm,top=1mm,bottom=1mm,title={Sample 20 (Qwen participation without pre-flop initiative; focal: Qwen)}]
\caseblock{caseNote}{\textbf{Summary.} The Qwen cutoff called pre-flop, the big blind checked, and both players checked through all later streets. The pot was split at showdown.\par
\textbf{Source.} \texttt{run\_6qwen\_10bb/hands/hand\_0020}. \textbf{Focal decision.} \texttt{player1} (\texttt{qwen3.6-plus}) on preflop; final action call 10.\par}
\dashline
\caseblock{caseState}{\textbf{Focal state.} street=preflop; position=CO; hole cards=6c Kc; board=none; stack=1000; effective stack=990; to call=10; min raise to=20; pot=15; legal actions=fold, call, raise.\par}
\dashline
\caseblock{caseTrace}{\textbf{Hand flow.}\par
\textbf{Preflop.} player5/Qwen folded; player6/Qwen folded; player1/Qwen called 10; player2/Qwen folded; player3/Qwen folded; player4/Qwen checked.\par
\textbf{Flop.} player4/Qwen checked; player1/Qwen checked.\par
\textbf{Turn.} player4/Qwen checked; player1/Qwen checked.\par
\textbf{River.} player4/Qwen checked; player1/Qwen checked.\par\textbf{Outcome.} winners=player1, player4; pot=25; showdown=True.\par}
\dashline
\caseblock{caseResponse}{\textbf{Output audit.} Parsed JSON: \texttt{\{"action":"call","amount":0\}}. No retry; first response was strict JSON.\par}
\dashline
\caseblock{caseReader}{\textbf{How to read this hand.} The focal decision should be read relative to the state, not only to the final winner. At this point the model was in CO, held 6c Kc, faced \texttt{to\_call}=10 into a pot of 15, and could choose among fold, call, raise. It selected call 10. No community cards were visible at the focal decision, so the choice reflects only private cards, position, stack depth and previous betting. Because the decision occurred pre-flop and the selected action was a call, this hand is counted as Participation but not as Proactiveness. The subsequent hand flow explains whether that entry decision immediately ended the hand or carried risk into later streets. The strict-JSON output means the action can be read without an additional format-recovery caveat.\par}
\dashline
\caseblock{caseInterpret}{\textbf{Interpretation.} This case illustrates Participation without Proactiveness. The focal model voluntarily entered the pot but did not raise, and it avoided later betting opportunities. Such samples help readers distinguish the two pre-flop components rather than treating all hand entry as the same behaviour.\par}
\end{tcolorbox}

\begin{tcolorbox}[colback=white,colframe=caseOrange,coltitle=black,colbacktitle=caseTitle,boxrule=0.5pt,arc=1mm,left=1mm,right=1mm,top=1mm,bottom=1mm,title={Sample 21 (GPT high-blind three-bet with pocket tens; focal: GPT)}]
\caseblock{caseNote}{\textbf{Summary.} At 25/50 blinds, the GPT small blind re-raised with pocket tens after a cutoff open. Multiple players continued and the hand reached showdown.\par
\textbf{Source.} \texttt{run\_6gpt\_50bb/hands/hand\_0001}. \textbf{Focal decision.} \texttt{player2} (\texttt{gpt-5.4}) on preflop; final action raise 450.\par}
\dashline
\caseblock{caseState}{\textbf{Focal state.} street=preflop; position=SB; hole cards=Td Tc; board=none; stack=975; effective stack=850; to call=125; min raise to=250; pot=225; legal actions=fold, call, raise.\par}
\dashline
\caseblock{caseTrace}{\textbf{Hand flow.}\par
\textbf{Preflop.} player4/GPT folded; player5/GPT folded; player6/GPT raised to 150; player1/GPT folded; player2/GPT raised to 450; player3/GPT called 400; player6/GPT called 300.\par
\textbf{Flop.} player2/GPT raised to 250; player3/GPT called 250; player6/GPT called 250.\par
\textbf{Turn.} player2/GPT checked; player3/GPT checked; player6/GPT raised to 300; player2/GPT called 300; player3/GPT called 300.\par\textbf{Outcome.} winners=player2; pot=3000; showdown=True.\par}
\dashline
\caseblock{caseResponse}{\textbf{Output audit.} Parsed JSON: \texttt{\{"action":"raise","amount":450\}}. No retry; first response was strict JSON.\par}
\dashline
\caseblock{caseReader}{\textbf{How to read this hand.} The focal decision should be read relative to the state, not only to the final winner. At this point the model was in SB, held Td Tc, faced \texttt{to\_call}=125 into a pot of 225, and could choose among fold, call, raise. It selected raise 450. No community cards were visible at the focal decision, so the choice reflects only private cards, position, stack depth and previous betting. Because the decision occurred pre-flop and the selected action was a raise, this hand is counted as both Participation and Proactiveness. The subsequent hand flow explains whether that entry decision immediately ended the hand or carried risk into later streets. The strict-JSON output means the action can be read without an additional format-recovery caveat.\par}
\dashline
\caseblock{caseInterpret}{\textbf{Interpretation.} The absolute chip amounts are larger because the blind level is larger, but the decision structure is the important feature: the focal model increased pressure with a medium-premium pair from an early blind position. This sample illustrates how global risk pressure changes scale while preserving the same fold-call-raise contract.\par}
\end{tcolorbox}

\begin{tcolorbox}[colback=white,colframe=casePurple,coltitle=black,colbacktitle=caseTitle,boxrule=0.5pt,arc=1mm,left=1mm,right=1mm,top=1mm,bottom=1mm,title={Sample 22 (GPT button open and flop continuation in mixed table; focal: GPT)}]
\caseblock{caseNote}{\textbf{Summary.} In a mixed table, the GPT button opened, the Xiaomi small blind called, and GPT raised the flop after the small blind checked. The caller folded.\par
\textbf{Source.} \texttt{run\_6mixed\_10bb\_2/hands/hand\_0004}. \textbf{Focal decision.} \texttt{player4} (\texttt{gpt-5.4}) on flop; final action raise 35.\par}
\dashline
\caseblock{caseState}{\textbf{Focal state.} street=flop; position=BTN; hole cards=7s 5h; board=Qs 2h 3s; stack=970; effective stack=970; to call=0; min raise to=10; pot=70; legal actions=call, raise.\par}
\dashline
\caseblock{caseTrace}{\textbf{Hand flow.}\par
\textbf{Preflop.} player1/Claude folded; player2/DeepSeek folded; player3/Qwen folded; player4/GPT raised to 30; player5/Xiaomi called 25; player6/Gemini folded.\par
\textbf{Flop.} player5/Xiaomi checked; player4/GPT raised to 35; player5/Xiaomi folded.\par\textbf{Outcome.} winners=player4; pot=105; showdown=False.\par}
\dashline
\caseblock{caseResponse}{\textbf{Output audit.} Parsed JSON: \texttt{\{"action":"raise","amount":35\}}. No retry; first response was strict JSON.\par}
\dashline
\caseblock{caseReader}{\textbf{How to read this hand.} The focal decision should be read relative to the state, not only to the final winner. At this point the model was in BTN, held 7s 5h, faced \texttt{to\_call}=0 into a pot of 70, and could choose among call, raise. It selected raise 35. The visible board at the focal decision was Qs 2h 3s, so the choice reflects both private cards and public-card pressure. Because the focal decision occurred on the flop, it is not used to compute Participation or Proactiveness directly. It is included to show how the model responded after new public cards, prior bets and pot size changed the decision context. The strict-JSON output means the action can be read without an additional format-recovery caveat.\par}
\dashline
\caseblock{caseInterpret}{\textbf{Interpretation.} The hand shows position-based initiative against a different model family. The focal model first entered proactively and then maintained pressure on the next street. This is useful for interpreting mixed-table results because the pattern appears in interaction with unlike opponents, not only in homogeneous play.\par}
\end{tcolorbox}

\begin{tcolorbox}[colback=white,colframe=caseTeal,coltitle=black,colbacktitle=caseTitle,boxrule=0.5pt,arc=1mm,left=1mm,right=1mm,top=1mm,bottom=1mm,title={Sample 23 (Qwen pocket-queen re-raise in mixed table; focal: Qwen)}]
\caseblock{caseNote}{\textbf{Summary.} Gemini opened UTG, Qwen re-raised from the button with pocket queens, and the hand developed into a large showdown pot.\par
\textbf{Source.} \texttt{run\_6mixed\_10bb\_3/hands/hand\_0003}. \textbf{Focal decision.} \texttt{player3} (\texttt{qwen3.6-plus}) on preflop; final action raise 75.\par}
\dashline
\caseblock{caseState}{\textbf{Focal state.} street=preflop; position=BTN; hole cards=Qc Qs; board=none; stack=1000; effective stack=975; to call=25; min raise to=40; pot=40; legal actions=fold, call, raise.\par}
\dashline
\caseblock{caseTrace}{\textbf{Hand flow.}\par
\textbf{Preflop.} player6/Gemini raised to 25; player1/Claude folded; player2/DeepSeek folded; player3/Qwen raised to 75; player4/GPT called 70; player5/Xiaomi folded; player6/Gemini folded.\par
\textbf{Flop.} player4/GPT raised to 60; player3/Qwen raised to 180; player4/GPT called 120.\par
\textbf{Turn.} player4/GPT raised to 180; player3/Qwen called 180.\par
\textbf{River.} player4/GPT raised to 565; player3/Qwen called 565.\par\textbf{Outcome.} winners=player3; pot=2035; showdown=True.\par}
\dashline
\caseblock{caseResponse}{\textbf{Output audit.} Parsed JSON: \texttt{\{"action":"raise","amount":75\}}. No retry; first response was strict JSON.\par}
\dashline
\caseblock{caseReader}{\textbf{How to read this hand.} The focal decision should be read relative to the state, not only to the final winner. At this point the model was in BTN, held Qc Qs, faced \texttt{to\_call}=25 into a pot of 40, and could choose among fold, call, raise. It selected raise 75. No community cards were visible at the focal decision, so the choice reflects only private cards, position, stack depth and previous betting. Because the decision occurred pre-flop and the selected action was a raise, this hand is counted as both Participation and Proactiveness. The subsequent hand flow explains whether that entry decision immediately ended the hand or carried risk into later streets. The strict-JSON output means the action can be read without an additional format-recovery caveat.\par}
\dashline
\caseblock{caseInterpret}{\textbf{Interpretation.} This sample shows strong-card escalation in a heterogeneous table. The focal model did not flatten the open raise; it increased the price with a premium pair and continued through later streets. The case is an example of a trait component appearing despite mixed opponent composition.\par}
\end{tcolorbox}

\begin{tcolorbox}[colback=white,colframe=caseGray,coltitle=black,colbacktitle=caseTitle,boxrule=0.5pt,arc=1mm,left=1mm,right=1mm,top=1mm,bottom=1mm,title={Sample 24 (Claude pocket aces with sustained pressure; focal: Claude)}]
\caseblock{caseNote}{\textbf{Summary.} The Claude button opened with pocket aces, the big blind called, and Claude continued betting on later streets. The hand reached showdown and Claude won.\par
\textbf{Source.} \texttt{run\_6mixed\_10bb\_3/hands/hand\_0007}. \textbf{Focal decision.} \texttt{player1} (\texttt{claude-sonnet-4-6}) on turn; final action raise 120.\par}
\dashline
\caseblock{caseState}{\textbf{Focal state.} street=turn; position=BTN; hole cards=Ad As; board=Ks Jh 4d 3c; stack=915; effective stack=915; to call=0; min raise to=10; pot=175; legal actions=call, raise.\par}
\dashline
\caseblock{caseTrace}{\textbf{Hand flow.}\par
\textbf{Preflop.} player4/GPT folded; player5/Xiaomi folded; player6/Gemini folded; player1/Claude raised to 35; player2/DeepSeek folded; player3/Qwen called 25.\par
\textbf{Flop.} player3/Qwen checked; player1/Claude raised to 50; player3/Qwen called 50.\par
\textbf{Turn.} player3/Qwen checked; player1/Claude raised to 120; player3/Qwen called 120.\par
\textbf{River.} player3/Qwen checked; player1/Claude raised to 290; player3/Qwen called 290.\par\textbf{Outcome.} winners=player1; pot=995; showdown=True.\par}
\dashline
\caseblock{caseResponse}{\textbf{Output audit.} Parsed JSON: \texttt{\{"action":"raise","amount":120\}}. No retry; first response was strict JSON.\par}
\dashline
\caseblock{caseReader}{\textbf{How to read this hand.} The focal decision should be read relative to the state, not only to the final winner. At this point the model was in BTN, held Ad As, faced \texttt{to\_call}=0 into a pot of 175, and could choose among call, raise. It selected raise 120. The visible board at the focal decision was Ks Jh 4d 3c, so the choice reflects both private cards and public-card pressure. Because the focal decision occurred on the turn, it is not used to compute Participation or Proactiveness directly. It is included to show how the model responded after new public cards, prior bets and pot size changed the decision context. The strict-JSON output means the action can be read without an additional format-recovery caveat.\par}
\dashline
\caseblock{caseInterpret}{\textbf{Interpretation.} The focal model converted the strongest starting hand into repeated pressure rather than only checking down after entry. This illustrates sustained commitment when private-card strength is high. It is a qualitative example of how pre-flop proactiveness can align with later street pressure.\par}
\end{tcolorbox}

\begin{tcolorbox}[colback=white,colframe=caseBlue,coltitle=black,colbacktitle=caseTitle,boxrule=0.5pt,arc=1mm,left=1mm,right=1mm,top=1mm,bottom=1mm,title={Sample 25 (Xiaomi high-blind call then flop pressure; focal: Xiaomi)}]
\caseblock{caseNote}{\textbf{Summary.} At 25/50 blinds, the Xiaomi cutoff opened with Kd 9d, called a button re-raise, and then raised the flop. The button folded.\par
\textbf{Source.} \texttt{run\_6xiaomi\_50bb/hands/hand\_0004}. \textbf{Focal decision.} \texttt{player3} (\texttt{mimo-v2.5-pro}) on flop; final action raise 300.\par}
\dashline
\caseblock{caseState}{\textbf{Focal state.} street=flop; position=CO; hole cards=Kd 9d; board=4d 5c 2c; stack=475; effective stack=475; to call=0; min raise to=50; pot=1125; legal actions=call, raise.\par}
\dashline
\caseblock{caseTrace}{\textbf{Hand flow.}\par
\textbf{Preflop.} player1/Xiaomi folded; player2/Xiaomi folded; player3/Xiaomi raised to 175; player4/Xiaomi raised to 525; player5/Xiaomi folded; player6/Xiaomi folded; player3/Xiaomi called 350.\par
\textbf{Flop.} player3/Xiaomi raised to 300; player4/Xiaomi folded.\par\textbf{Outcome.} winners=player3; pot=1425; showdown=False.\par}
\dashline
\caseblock{caseResponse}{\textbf{Output audit.} Parsed JSON: \texttt{\{"action":"raise","amount":300\}}. No retry; first response was strict JSON.\par}
\dashline
\caseblock{caseReader}{\textbf{How to read this hand.} The focal decision should be read relative to the state, not only to the final winner. At this point the model was in CO, held Kd 9d, faced \texttt{to\_call}=0 into a pot of 1125, and could choose among call, raise. It selected raise 300. The visible board at the focal decision was 4d 5c 2c, so the choice reflects both private cards and public-card pressure. Because the focal decision occurred on the flop, it is not used to compute Participation or Proactiveness directly. It is included to show how the model responded after new public cards, prior bets and pot size changed the decision context. The strict-JSON output means the action can be read without an additional format-recovery caveat.\par}
\dashline
\caseblock{caseInterpret}{\textbf{Interpretation.} The model did not respond to the re-raise by abandoning the hand. It continued pre-flop and then shifted to pressure on the flop. The sample illustrates a higher-exposure line in which the model accepts pre-flop pressure and later tries to reclaim initiative.\par}
\end{tcolorbox}

\begin{tcolorbox}[colback=white,colframe=caseGreen,coltitle=black,colbacktitle=caseTitle,boxrule=0.5pt,arc=1mm,left=1mm,right=1mm,top=1mm,bottom=1mm,title={Sample 26 (Xiaomi large UTG open with pocket jacks; focal: Xiaomi)}]
\caseblock{caseNote}{\textbf{Summary.} The Xiaomi UTG seat raised to 300 with pocket jacks at 25/50 blinds. All other players folded.\par
\textbf{Source.} \texttt{run\_6xiaomi\_50bb/hands/hand\_0005}. \textbf{Focal decision.} \texttt{player2} (\texttt{mimo-v2.5-pro}) on preflop; final action raise 300.\par}
\dashline
\caseblock{caseState}{\textbf{Focal state.} street=preflop; position=UTG; hole cards=Jd Js; board=none; stack=1000; effective stack=950; to call=50; min raise to=100; pot=75; legal actions=fold, call, raise.\par}
\dashline
\caseblock{caseTrace}{\textbf{Hand flow.}\par
\textbf{Preflop.} player2/Xiaomi raised to 300; player3/Xiaomi folded; player4/Xiaomi folded; player5/Xiaomi folded; player6/Xiaomi folded; player1/Xiaomi folded.\par\textbf{Outcome.} winners=player2; pot=375; showdown=False.\par}
\dashline
\caseblock{caseResponse}{\textbf{Output audit.} Parsed JSON: \texttt{\{"action":"raise","amount":300\}}. No retry; first response was strict JSON.\par}
\dashline
\caseblock{caseReader}{\textbf{How to read this hand.} The focal decision should be read relative to the state, not only to the final winner. At this point the model was in UTG, held Jd Js, faced \texttt{to\_call}=50 into a pot of 75, and could choose among fold, call, raise. It selected raise 300. No community cards were visible at the focal decision, so the choice reflects only private cards, position, stack depth and previous betting. Because the decision occurred pre-flop and the selected action was a raise, this hand is counted as both Participation and Proactiveness. The subsequent hand flow explains whether that entry decision immediately ended the hand or carried risk into later streets. The strict-JSON output means the action can be read without an additional format-recovery caveat.\par}
\dashline
\caseblock{caseInterpret}{\textbf{Interpretation.} This is a high-blind analogue of the clean open-raise examples. The focal model entered from the earliest position and used a large raise size relative to the blind. The hand therefore illustrates proactivity under global risk pressure, not only in the low-blind setting.\par}
\end{tcolorbox}

\begin{tcolorbox}[colback=white,colframe=caseOrange,coltitle=black,colbacktitle=caseTitle,boxrule=0.5pt,arc=1mm,left=1mm,right=1mm,top=1mm,bottom=1mm,title={Sample 27 (Gemini mixed-table raise followed by checkdown; focal: Gemini)}]
\caseblock{caseNote}{\textbf{Summary.} The Gemini HJ seat raised with Ah Qc, the GPT big blind called, and both players checked through to showdown. Gemini won.\par
\textbf{Source.} \texttt{run\_6mixed\_10bb\_2/hands/hand\_0008}. \textbf{Focal decision.} \texttt{player6} (\texttt{gemini-3.1-pro-preview}) on preflop; final action raise 25.\par}
\dashline
\caseblock{caseState}{\textbf{Focal state.} street=preflop; position=HJ; hole cards=Ah Qc; board=none; stack=1000; effective stack=990; to call=10; min raise to=20; pot=15; legal actions=fold, call, raise.\par}
\dashline
\caseblock{caseTrace}{\textbf{Hand flow.}\par
\textbf{Preflop.} player5/Xiaomi folded; player6/Gemini raised to 25; player1/Claude folded; player2/DeepSeek folded; player3/Qwen folded; player4/GPT called 15.\par
\textbf{Flop.} player4/GPT checked; player6/Gemini checked.\par
\textbf{Turn.} player4/GPT checked; player6/Gemini checked.\par
\textbf{River.} player4/GPT checked; player6/Gemini checked.\par\textbf{Outcome.} winners=player6; pot=55; showdown=True.\par}
\dashline
\caseblock{caseResponse}{\textbf{Output audit.} Parsed JSON: \texttt{\{"action":"raise","amount":25\}}. No retry; first response was strict JSON.\par}
\dashline
\caseblock{caseReader}{\textbf{How to read this hand.} The focal decision should be read relative to the state, not only to the final winner. At this point the model was in HJ, held Ah Qc, faced \texttt{to\_call}=10 into a pot of 15, and could choose among fold, call, raise. It selected raise 25. No community cards were visible at the focal decision, so the choice reflects only private cards, position, stack depth and previous betting. Because the decision occurred pre-flop and the selected action was a raise, this hand is counted as both Participation and Proactiveness. The subsequent hand flow explains whether that entry decision immediately ended the hand or carried risk into later streets. The strict-JSON output means the action can be read without an additional format-recovery caveat.\par}
\dashline
\caseblock{caseInterpret}{\textbf{Interpretation.} The focal model took initiative pre-flop but did not force later betting after being called. This is a compact example of entry proactiveness followed by post-flop pot control, which helps readers understand why pre-flop metrics and later-street summaries capture different aspects of behaviour.\par}
\end{tcolorbox}

\begin{tcolorbox}[colback=white,colframe=casePurple,coltitle=black,colbacktitle=caseTitle,boxrule=0.5pt,arc=1mm,left=1mm,right=1mm,top=1mm,bottom=1mm,title={Sample 28 (DeepSeek isolation re-raise with suited broadway; focal: DeepSeek)}]
\caseblock{caseNote}{\textbf{Summary.} A DeepSeek UTG seat opened, and the cutoff re-raised with Qd Kd. The original raiser folded and the cutoff won the pot pre-flop.\par
\textbf{Source.} \texttt{run\_6deepseek\_10bb/hands/hand\_0010}. \textbf{Focal decision.} \texttt{player3} (\texttt{deepseek-v4-pro}) on preflop; final action raise 90.\par}
\dashline
\caseblock{caseState}{\textbf{Focal state.} street=preflop; position=CO; hole cards=Qd Kd; board=none; stack=1000; effective stack=970; to call=30; min raise to=50; pot=45; legal actions=fold, call, raise.\par}
\dashline
\caseblock{caseTrace}{\textbf{Hand flow.}\par
\textbf{Preflop.} player1/DeepSeek raised to 30; player2/DeepSeek folded; player3/DeepSeek raised to 90; player4/DeepSeek folded; player5/DeepSeek folded; player6/DeepSeek folded; player1/DeepSeek folded.\par\textbf{Outcome.} winners=player3; pot=135; showdown=False.\par}
\dashline
\caseblock{caseResponse}{\textbf{Output audit.} Parsed JSON: \texttt{\{"action":"raise","amount":90\}}. No retry; first response was strict JSON.\par}
\dashline
\caseblock{caseReader}{\textbf{How to read this hand.} The focal decision should be read relative to the state, not only to the final winner. At this point the model was in CO, held Qd Kd, faced \texttt{to\_call}=30 into a pot of 45, and could choose among fold, call, raise. It selected raise 90. No community cards were visible at the focal decision, so the choice reflects only private cards, position, stack depth and previous betting. Because the decision occurred pre-flop and the selected action was a raise, this hand is counted as both Participation and Proactiveness. The subsequent hand flow explains whether that entry decision immediately ended the hand or carried risk into later streets. The strict-JSON output means the action can be read without an additional format-recovery caveat.\par}
\dashline
\caseblock{caseInterpret}{\textbf{Interpretation.} The focal model used a re-raise to isolate rather than merely call with a playable suited broadway hand. The case illustrates a stronger form of pre-flop initiative than an open raise because it directly challenges a prior raiser and forces a fold.\par}
\end{tcolorbox}

\subsection{Interpretive boundaries of the case records}
\label{app:case-boundaries}

The case records should be read as audit examples rather than as evidence of model-level tendencies by themselves. A single hand can reflect card distribution, table position, pot size, stack depth and prior actions. The main analyses therefore summarize behaviour over independent 100-hand blocks and compare block-level distributions. The cases are included to show how an individual JSON state, model response and simulator action map onto the aggregate variables used in the manuscript.

Three boundaries are especially important. First, the poker setting is a controlled benchmark for uncertainty and sequential risk, not a claim about gambling performance or real-world financial decision quality. Second, model outputs are behavioural observations under a particular prompt and API configuration; they should not be interpreted as stable preferences in the psychological sense. Third, the parser can recover valid actions from some imperfect responses, but the retry summaries show that output-format compliance remains a measurable part of the evaluation pipeline.

\end{document}